\documentclass[poms,final,unblindrev]{poms1_V1} % Use this for listing out authors' information submissions

\OneAndAHalfSpacedXI % % default line spacing !!!!!!!!!!!!!!!!!!!!!!!
\usepackage{graphicx}
\usepackage{subfigure, epsfig}
\usepackage{natbib}
\usepackage{amsmath}
	\usepackage{graphicx}
	\usepackage{enumerate}
	\usepackage{natbib} %comment out if you do not have the package
	\usepackage{url} % not crucial - just used below for the URL
	\usepackage{hyperref}       % hyperlinks
    \usepackage{url}            % simple URL typesetting
    \usepackage{booktabs}       % professional-quality tables
    \usepackage{amsfonts}       % blackboard math symbols
    \usepackage{nicefrac}       % compact symbols for 1/2, etc.
    \usepackage{microtype}      % microtypography
    \usepackage{graphicx}
    \usepackage[space]{grffile}
    \usepackage{latexsym}
    \usepackage{textcomp}
    \usepackage{longtable}
    \usepackage{tabulary}
    \usepackage{booktabs,array,multirow}
    \usepackage{amsfonts,amsmath,amssymb}
    \usepackage{url}
    \usepackage{mathtools}
    \usepackage{bm}
    \usepackage[mathscr]{euscript}
    \usepackage{enumitem}
    \usepackage{xcolor}
    \usepackage{multirow}
    \usepackage{booktabs}
    \usepackage{microtype}
    \usepackage{fix-cm}
    \usepackage{colortbl}
    \usepackage{tikz}
    \usepackage{pbox}
    \usepackage{xcolor}
    \usepackage{etoolbox}
    \usepackage{appendix}
    \usepackage{scalerel}
    \usepackage[english]{babel}
    \usepackage{gensymb}
    \usepackage{graphicx}
\usepackage{epstopdf}
\usepackage{moreverb}
\usepackage{algorithm}
\usepackage{algpseudocode}

    		\DeclareMathOperator{\pr}{\mathbb P}
\DeclareMathOperator{\E}{\mathbb E}

 \bibpunct[, ]{(}{)}{,}{a}{}{,}%
 \def\bibfont{\small}%
\TheoremsNumberedThrough     % Preferred (Theorem 1, Lemma 1, Theorem 2)
\ECRepeatTheorems

\newcommand{\Real}{\mathbb R}

\newcommand{\CalN}{\mathcal N}
\newcommand{\CalO}{\mathcal O}
\newcommand{\CalG}{\mathcal G}

\newcommand{\rmI}{\boldsymbol{\mathrm I}}
\newcommand{\BFx}{\boldsymbol{x}}
\newcommand{\BFX}{\boldsymbol{X}}
\newcommand{\BFy}{\boldsymbol{y}}
\newcommand{\BFP}{\boldsymbol{P}}
\newcommand{\BFb}{\boldsymbol{b}}
\newcommand{\BFomega}{\boldsymbol{\omega}}
\newcommand{\BFB}{\boldsymbol{B}}
\newcommand{\BFY}{\boldsymbol{Y}}
\newcommand{\BFe}{\boldsymbol{e}}
\newcommand{\BFK}{\boldsymbol{K}}
\newcommand{\BFA}{\boldsymbol{A}}

\newcommand{\BFv}{\boldsymbol{v}}
\newcommand{\BFM}{\boldsymbol{M}}
\newcommand{\BFH}{\boldsymbol{H}}
\newcommand{\BFPhi}{\boldsymbol{\Phi}}
\newcommand{\BFphi}{\boldsymbol{\phi}}

\newcommand{\BFPsi}{\boldsymbol{\Psi}}
\newcommand{\BFzero}{\boldsymbol{0}}
\newcommand{\BFS}{\boldsymbol{S}}
\newcommand{\BFthete}{\boldsymbol{\theta}}

\newcommand{\specialcell}[2][c]{%
  \begin{tabular}[#1]{@{}c@{}}#2\end{tabular}}
\EquationsNumberedThrough    % Default: (1), (2), ...
\begin{document}
%%%%%%%%%%%%%%%%

%\setcounter{page}{1}
%\DOI{10.1287/xxxx.0000.0000}%

\RUNAUTHOR{Zou, Chen, Zhou }

\RUNTITLE{ Representing Additive Gaussian Processes by Sparse Matrices}

\TITLE{ Representing Additive Gaussian Processes by \\Sparse Matrices}

\ARTICLEAUTHORS{%
\AUTHOR{Lu Zou}
\AFF{Information Hub, The Hong Kong University of Science and Technology (Guangzhou) }%\EMAIL{jdoe@operations.edu}} %, \URL{}}
\AUTHOR{Haoyuan Chen}
\AFF{Department of Industrial and Systems Engineering  }%\EMAIL{}}
\AUTHOR{Liang Ding}
\AFF{School of Data Science, Fudan University,
\EMAIL{liang\_ding@fudan.edu.cn}}
} % end of the block

\ABSTRACT{%
Among generalized additive models, additive Mat\'ern Gaussian Processes (GPs) are one of the most popular for scalable high-dimensional problems. Thanks to their additive structure and stochastic differential equation representation, back-fitting-based algorithms can reduce the time complexity of computing the posterior mean from $\CalO(n^3)$ to $\CalO(n\log n)$ time where $n$ is the data size. However, generalizing these algorithms to efficiently compute the posterior variance and maximum log-likelihood  remains an open problem. In this study, we demonstrate that for Additive Mat\'ern GPs, not only the posterior mean, but also the posterior variance, log-likelihood, and gradient of these three functions can be represented by formulas involving only sparse matrices and sparse vectors. We show how to use these sparse formulas to generalize back-fitting-based algorithms to efficiently compute the posterior mean, posterior variance, log-likelihood, and gradient of these three functions for additive GPs, all in $\CalO(n \log n)$ time.  We apply our algorithms to Bayesian optimization and propose efficient algorithms for posterior updates, hyperparameters learning, and computations of the acquisition function and its gradient in Bayesian optimization. Given the posterior, our algorithms significantly reduce the time complexity of computing the acquisition function and its gradient from $\CalO(n^2)$ to $\CalO(\log n)$ for general learning rate, and even to $\CalO(1)$ for small learning rate.
% Enter your abstract
}%

%Use this for final submission
%\HISTORY{This paper was first submitted on January 1, 2021 and has been with the authors for 3 months for 2 revisions.}

\maketitle
%%%%%%%%%%%%%%%%%%%%%%%%%%%%%%%%%%%%%%%%%%%%%%%%%%%%%%%%%%%%%%%%%%%%%%

% Samples of sectioning (and labeling) in POMS
% NOTE: (1) \section and \subsection do NOT end with a period
%       (2) \subsubsection and lower need end punctuation
%       (3) capitalization is as shown (title style).
%
%\section{Introduction.}\label{intro} %%1.
%\subsection{Duality and the Classical EOQ Problem.}\label{class-EOQ} %% 1.1.
%\subsection{Outline.}\label{outline1} %% 1.2.
%\subsubsection{Cyclic Schedules for the General Deterministic SMDP.}
%  \label{cyclic-schedules} %% 1.2.1
%\section{Problem Description.}\label{problemdescription} %% 2.

% Text of your paper here

\section{Introduction}\label{sec:intro}
Among generalized additive model \cite{hastie2017generalized}, additive GPs have gained popularity as priors in scalable problems due to their ability to accurately estimate targets with intrinsic low-dimensional structures \citep{KandasamySchneiderPoczos15,RollandScarlettBogunovicCevher18,delbridge2020randomly}. They have been applied to many field, such as Bayesian optimization \citep{KandasamySchneiderPoczos15,wang2017batched}, simulation metamodeling\citep{chen2022projection}, and bandits \cite{cai2022stochastic}. 
 However, for large data sizes, computing additive GP Bayesian optimization becomes inefficient, requiring $\CalO(n^3)$ time for posterior updates and $\CalO(n^2)$ time for prediction given the posterior. While backfitting-based algorithms and the stochastic differential equation representation of additive Mat\'ern GPs have enabled efficient computation of the posterior mean \citep{gilboa2013scaling,saatcci2012scalable}, generalizing these algorithms to efficiently compute the posterior variance and maximum log-likelihood remains an open problem.

The posterior variance and log-likelihood are crucial in the context of additive GPs. For instance, in Bayesian optimization, finding the next sampling point involves learning the hyperparameters by maximizing the log-likelihood and computing the acquisition function's maximizer, which is a function of the posterior mean and variance. Gradient-based methods are typically employed for this purpose, where the gradient of the log-likelihood and acquisition function is iteratively computed to update information about the maximizer. However, as these computations involve large matrix inverses, determinants, and traces, each iteration in the search for the maximizer takes $\CalO(n^3)$ time if direct computation is used. This high computational cost is impractical and significantly increases the time required to find the next sampling point.

In this study, we generalize back-fitting algorithms to compute the posterior and log-likelihood of additive Mat\'ern GPs. Specifically, we demonstrate that the posterior mean, posterior variance, and log-likelihood of additive GPs and the gradient of these functions can all be expressed as summations of functions known as Kernel Packets (KPs) \citep{chen2022kernel}, which have compact and almost mutually disjoint supports. This allows us to reformulate the posterior, log-likelihood, and their gradient as multiplications of sparse matrices and vectors, which in turn enables us to generalize back-fitting algorithms for updating the posterior, learning hyperparameters, and computing the gradient of these functions with much greater efficiency than previous methods. In particular, our algorithms reduce the time complexity of computing the posterior mean, posterior variance, log-likelihood and their gradient from $\CalO(n^3)$ to $\CalO(n\log n)$.  When applying our algorithms to Bayesian optimization, time complexity of computing  the acquisition function and its gradient can be reduced from $\CalO(n^2)$ to $\CalO(\log n)$ and even to $\CalO(1)$ if the learning rate is small enough, given the posterior. These advantages make our approach a significant improvement over existing methods and greatly enhance the computational efficiency of additive GP Bayesian optimization.

\section{Backgrounds}
In this section, we provide a brief introduction to GP regression and review some existing methods in Bayesian optimization based on GPs.

\subsection{Gaussian Processes}
GP is a popular Bayesian method for nonparametric regression, which allows the specification of a prior distribution over continuous functions via a Gaussian process.  A comprehensive treatment of GPs can be found in  \cite{RasmussenWilliams06}. A GP is a distribution on function $\CalG(\cdot)$ over an input space $\bold{U}$ such
that the distribution of $\CalG$ on any size-$n$ set of  input points $\BFX=\{\BFx_i\}_{i=1}^n\subset \bold{U}$  is described by a multivariate Gaussian density over the associated targets, i.e.,
\[\pr\left(\CalG(\BFx_1),\cdots,\CalG(\BFx_n)\right)=\CalN(m(\BFX),k(\BFX,\BFX))\]
where $m(\BFX)\in\Real^n$ is an $n$-vector whose $i$-entry equals the value of a mean function $m$ on point $\BFx_i$ and $k(\BFX,\BFX)=[k(\BFx_i,\BFx_j)]_{i,j=1}^n\in\Real^{n\times n}$ is an $n$-by-$n$ matrix whose $(i,j)$-entry equal the value of a positive definite kernel $k$ on $\BFx_i$ and $\BFx_j$ \citep{wendland2004scattered}. Accordingly, a GP can be characterized by the mean function $m:\bold{U}\to \Real$ and the kernel function $k :\bold{U}\times\bold{U}\to \Real$. Also , the mean function is often set as $0$ when we have limited knowledge of the true function. Therefore, we can use kernel $k$ only to determine a GP.

We consider the case where what we observe is a noisy version of the underlying
function, $\BFy_i=\CalG(\BFx_i)+\varepsilon_i$, where $\varepsilon_i\sim\CalN(0,\sigma_y^2)$ is i.i.d. Gaussian distributed error. Then, we can use standard identities of the multivariate Gaussian
distribution to show that, conditioned on data $(\BFX,\BFY)=\{(\BFx_i,\BFy_i)\}_{i=1}^n$, the posterior distribution at any point $\BFx^*$ also follows a Gaussian distribution: $\CalG(\BFx^*) \vert \BFX,\BFY\sim\CalN(\mu_n(\BFx^*),{s}_n(\BFx^*))$, where
\begin{align}
&\mu_n(\BFx^*)=k(\BFx^*,\BFX)\left[k(\BFX,\BFX)+\sigma_y^2\rmI_n\right]^{-1}\BFY\nonumber\\
    &s_n(\BFx^*)=k(\BFx^*,\BFx^*)-k(\BFx^*,\BFX)\left[k(\BFX,\BFX)+\sigma_y^2\rmI_n\right]^{-1}k(\BFX,\BFx^*)\label{eq:GP_posterior}
\end{align}
In the case kernel function $k(\cdot,\cdot\vert\BFthete)$ is parametrized by hyperparameters $\BFthete$, we can optimize the following negative log marginal likelihood function with respective to $\BFthete$:
\begin{equation}
    {l}(\BFthete)\propto-\BFY^T\left[k(\BFX,\BFX\vert {\BFthete})+\sigma_y^2\rmI_n\right]^{-1}\BFY-\log \vert k(\BFX,\BFX\vert {\BFthete})+\sigma_y^2\rmI_n\vert \label{eq:loglik}
\end{equation}

In order to have accurate prediction, we first need to search the maximizer of \eqref{eq:loglik}: $\BFthete^*=\arg\max_{\BFthete}l(\BFthete)$, which involves computing  the inverse matrix $\left[k(\BFX,\BFX\vert {\BFthete})+\sigma_y^2\rmI_n\right]^{-1}$ and the determinant $\vert k(\BFX,\BFX\vert {\BFthete})+\sigma_y^2\rmI_n\vert $. Then, we substitute the estimated $\BFthete^*$ into \eqref{eq:GP_posterior} and compute the 
  inverse matrix $\left[k(\BFX,\BFX\vert {\BFthete^*})+\sigma_y^2\rmI_n\right]^{-1}$ and vector $\BFb_{\BFY}=\left[k(\BFX,\BFX\vert {\BFthete}^*)+\sigma_y^2\rmI_n\right]^{-1}\BFY$  associated  to $\BFthete^*$. These operations require $\CalO(n^3)$ time complexity in general. At last, for a new predictive point $\BFx^*$, we can compute the posterior $\CalN(\mu_n(\BFx^*),s_n(\BFx^*))$, which involves matrix-vector multiplications $\left[k(\BFX,\BFX\vert {\BFthete})+\sigma_y^2\rmI_n\right]^{-1}k(\BFX,\BFx^*)$ and vector multiplication $k(\BFx^*,\BFX)\BFb_{\BFY}$.  These matrix multiplications require $\CalO(n^2)$ and $\CalO(n)$ time, respectively, given  $\left[k(\BFX,\BFX\vert {\BFthete})+\sigma_y^2\rmI_n\right]^{-1}$ and $\BFb_{\BFY}$ are known.

\subsection{Bayesian Optimization}
In Bayesian optimization, we treat the unknown function $\CalG$ as GP and evaluate it over a set of input points, denoted by $\BFx_1,\cdots,\BFx_n$. We call them the design points, because these points can be chosen according to the actual requirement. There are two categories of strategies to choose design points. Firstly, We can choose all the points before we evaluate $\CalG$ at any of them. Such a design set is call a fixed design. An alternative strategy is called sequential sampling, in which the design points are not fully determined at the beginning. Instead, points are added sequentially, guided by the information from the previous input points and the corresponding acquired function values. An instance algorithm defines a sequential sampling scheme which determines the next input point $\BFx_{n+1}$ by optimizing an acquisition function $\max_{\BFx\in U}A(x,\BFX,\BFY)$ where $\BFX=\{\BFx_i\}_{i=1}^n$ consists of the previous selections and $\BFY=\{\BFy_i\}_{i=1}^n$ are the noisy observations on $\BFX$ as defined before. A general Bayesian optimization procedure under sequential sampling scheme can be summarized as the following algorithm:
\begin{algorithm}
\caption{Bayesian Optimization}\label{alg:BayesOpt}
\hspace*{\algorithmicindent} \textbf{Input}: Gaussian prior $\CalG$, initial data $(\BFX_0,\BFY_0)$, and sampling budget $N$ \\
 \hspace*{\algorithmicindent} \textbf{Output}: maximizer of the posterior mean: $\BFx_{\text{max}}=\arg\max_{\BFx}\mu_n(\BFx)$
\begin{algorithmic}
\State $(\BFX,\BFY)\leftarrow (\BFX_0,\BFY_0)$
\For{$n= 1$ to $N$}
\State search $\BFx_n=\arg\max_{\BFx}A(\BFx,\BFX,\BFY)$
\State Sample $\BFy_n=\CalG(\BFx_n)+\varepsilon_n$
\State $(\BFX,\BFY)\leftarrow (\BFx_n,\BFy_n)$
\State Update posterior and hyperparameters $\BFthete$ of $\CalG$ conditioned on $(\BFX,\BFY)$
\EndFor
\end{algorithmic}
\end{algorithm}
The acquisition functions evaluate the “goodness” of a point $\BFx$ based on the posterior distribution defined by $\mu_n(\BFx),s_n(\BFx)$ at some hyperparameters $\BFthete^*$. The following two acquisition functions are among the most popular: 
\begin{enumerate}
    \item Gaussian process upper confidence bound (GP-UCB) \citep{SrinivasKrauseKakadeSeeger10} chooses the point at which the upper confidence bound is currently the highest in the $n$-th iteration. Its acquisition function is called the upper confidence bound and defined as $A(\BFx,\BFX,\BFY)=\mu_n(\BFx)+\beta_n\sqrt{s_n(\BFx)}$ where $\beta_n$ is the bandwidth hyperparameter. 
    \item \emph{Expected Improvement} (EI) \citep{jones1998} evaluates the expected amount of improvement in the objective function and aims at selecting the point that maximizing the improvement. Its acquisition is called the expected improvement and  defined as  $A(x,\BFX,\BFY)=\E\left[(\CalG(\BFx)-\max_{\BFy\in\BFY} \BFy)^+\vert\BFX,\BFY\right]$ where $(f)^+:=\max\{0,f\}$ denotes the non-negative part of $f$.
\end{enumerate}

In order to search the next sampling point, we must compute gradient of the acquisition function with respective to $\BFx$: $\partial A/\partial \BFx$. This operation involves computing the gradient of the posterior mean 
\[\frac{\partial \mu_n(\BFx)}{\partial \BFx}=\sum_{i=1}^n\frac{\partial k(\BFx_i,\BFx)}{\partial \BFx}\BFb_i, \quad \text{where} \quad\BFb=\left[k(\BFX,\BFX)+\sigma_y^2\rmI_n\right]^{-1}\BFY,\]
and the gradient of the posterior variance
\[\frac{\partial s_n(\BFx)}{\partial \BFx}=\frac{\partial k(\BFx,\BFx)}{\partial \BFx}-2\sum_{i,j=1}^n\frac{\partial k(\BFx_i,\BFx)}{\partial \BFx}\BFM_{i,j}k(\BFx_j,\BFx), \] 
where $\BFM=\left[k(\BFX,\BFX)+\sigma_y^2\rmI_n\right]^{-1}.$
The above equations demonstrate that computing the gradient of the acquisition function $A(\BFx, \BFX, \BFY)$ requires a minimum of $\CalO(n^2)$ time, assuming the inverse of posterior covariance matrix is known. In many large-scale problems, hundreds of thousands of gradient ascent steps may be necessary to search for the next sampling point, each of which also requires $\CalO(n^2)$ time complexity. As the number of iterations $n$ increases, this time complexity can make the runtime of Bayesian optimization excessively long.

\section{Additive Gaussian Processes}
A $D$-dimensional additive GP can be seen as summation of $D$ one-dimensional GPs. In specific, we use the following model to describe the generation of oberavation $\BFY$:
\begin{equation}
    \label{eq:additive_GP}
    \BFy_i=\sum_{d=1}^D\CalG_d(\BFx_{i,d}\vert\BFthete_d)+\varepsilon_i
\end{equation}
where $\CalG_d$ is a one-dimensional zero-mean GP characterized by kernel $k_d$, $\BFthete_d$ is the hyperparameters for kernel $k_d$, and $\BFx_{i,d}$ is the $d$-th entry of the $D$-dimensional input point $\BFx_i$. Given data $(\BFX,\BFY)$, we first use the following theorems to rewrite the posterior and log likelihood in forms that consist  of  one-dimensional GPs covariance matrices.
\begin{theorem}
\label{thm:posterior_block}
Conditioned on data $(\BFX,\BFY)$, the posterior distribution at any point $\BFx^*$ of an additive GP \eqref{eq:additive_GP} follows a Gaussian distribution: $\CalG(\BFx^*)\vert\BFX,\BFY\sim\CalN(\mu_n(\BFx^*),s_n(\BFx^*))$, where
\begin{align}
    &\mu_n(\BFx^*)=\boldsymbol{1}^T\boldsymbol{\gamma}_{\BFx^*}^T\BFK^{-1}[\BFK^{-1}+\frac{1}{\sigma_y^2}\BFS\BFS^T]^{-1}\BFS(\frac{1}{\sigma_y^2}\rmI_n)\BFY\nonumber \\
    &s_n(\BFx^*)=\sum_{d=1}^Dk_d(x^*_d,\BFx^*_d)-\boldsymbol{1}^T\boldsymbol{\gamma_{\BFx^*}}^T\BFK^{-1}\boldsymbol{\gamma_{\BFx^*}}\boldsymbol{1}\nonumber\\
    &\quad \quad\quad\quad+\boldsymbol{1}^T\boldsymbol{\gamma}_{\BFx^*}^T\BFK^{-1}[\BFK^{-1}+\frac{1}{\sigma_y^2}\BFS\BFS^T]^{-1}\BFK^{-1}\boldsymbol{\gamma}_{\BFx^*}\boldsymbol{1}\label{eq:posterior_block}
\end{align}
and
\begin{align*}
    &\BFK=\begin{bmatrix}
k_1(\BFX_1,\BFX_1) & & &\\
&k_2(\BFX_2,\BFX_2)& &\\
& &\ddots &\\
& & & k_D(\BFX_D,\BFX_D)
\end{bmatrix}\in\Real^{Dn\times Dn}\\
&\boldsymbol{\gamma}_{\BFx^*}=\begin{bmatrix}
k_1(\BFX_1,x^*_1) & & &\\
&k_2(\BFX_2,x^*_2)& &\\
& &\ddots &\\
& & & k_D(\BFX_D,x^*_D)
\end{bmatrix}\in\Real^{Dn\times D},
\end{align*}
$\BFX_d=\{\BFx_{i,d}\}_{i=1}^n$ denotes the $d$-th dimension of all data points $\{\BFx_i\}$, $\BFS=[\rmI_n;\rmI_n;\cdots;\rmI_n]\in\Real^{Dn\times n}$, and $\bold{1}=[1;1;\cdots;1]$ denotes the vector with all entries equal 1.
\end{theorem}

\begin{theorem}\label{thm:add_likelihood}
    The likelihood  can be written as 
    \begin{align}
        &l(\BFthete)\propto -\frac{\BFY}{D}^T\BFS^T\left(\BFK^{-1}_{\BFthete}-\BFK^{-1}_{\BFthete}[\BFK^{-1}_{\BFthete}+\sigma_y^{-2}\BFS\BFS^T]^{-1}\BFK^{-1}_{\BFthete}\right)\BFS\frac{\BFY}{D} \nonumber\\
        &\quad -\log \vert\left(\BFK^{-1}_{\BFthete}+\sigma_y^{-2}\BFS\BFS^T\right)\vert+\log \vert\BFK_{\BFthete}^{-1}\vert-2n\log\sigma_y \label{eq:add_likelihood} 
        \end{align}
        and its gradient  with respective to $\BFthete_d$ can be written as
        \begin{align}
        &\frac{\partial l}{\partial \BFthete_d}\propto\BFY^TR\big[\partial _{ \BFthete_d}k_d(\BFX_d,\BFX_d\vert\BFthete_d)\big]R\BFY-\text{Trace}\left[R\big[\partial _{ \BFthete_d}k_d(\BFX_d,\BFX_d\vert\BFthete_d)\big]\right]\label{eq:add_likelihood_gradient}
    \end{align}
    where $\BFK_{\BFthete}$ is the $Dn$-by-$Dn$ covariance in \eqref{eq:posterior_block} induced by kernel $k(\cdot,\cdot\vert\BFthete)$ and
    \[R=[\BFS^T\BFK_{\BFthete}\BFS+\sigma_y^2\rmI_n]^{-1}=\frac{1}{\sigma_y^2}\rmI_n-\frac{1}{\sigma_y^2}\BFS^T[\BFK_{\BFthete}^{-1}+\frac{1}{\sigma_y^2}\BFS\BFS^T]^{-1}\BFS\frac{1}{\sigma_y^2}.\]
\end{theorem}

Proofs of Theorem \ref{thm:posterior_block} and Theorem \ref{thm:add_likelihood} are left in Appendix. The block matrices associated with one-dimensional GPs in Theorem \ref{thm:posterior_block} and Theorem \ref{thm:add_likelihood} demonstrate that the computation of an additive GP can be decomposed into computations of D one-dimensional GPs. If there is a sparse representation of the covariance matrix $k_d(\BFX_d,\BFX_d)$ for each $d$, we can accelerate the computation of additive GP.

\section{Sparse Factorization }
In this subsection, we show the sparse formulations of one-dimensional Mat\'tern kernels. A one-dimensional Mat\'ern kernel function \citep{wendland2004scattered} is written as:
\begin{equation}
    \label{eq:Matern-1d}
     k(x,x')=\frac{2^{1-\nu}}{\Gamma(\nu)}\left(\sqrt{2\nu}{\omega}{\lvert x-x'\rvert}\right)^{\nu}K_{\nu}\left(\sqrt{2\nu}{\omega}{\lvert x-x'\rvert}\right),
\end{equation}
for any $x,x'\in\Real$, where $\nu>0$ is the smoothness parameter, $\omega>0$ is the scale and $K_\nu$ is the modified Bessel function of the second kind. The smoothness parameter $\nu$ governs the smoothness of the GP; the scale parameter $\omega$  determines the spread of the covariance. Mat\'ern covariances  are widely used because of its great flexibility. Therefore, the hyperparameters of an additive Mat\'ern-$\nu$ is the scale parameters of each dimension $\boldsymbol{\omega}=\{\omega_d\}_{d=1}^D$.

In particular, when the smoothness parameter equals half-integer, i.e., $\nu=1/2,3/2,5/2,\cdots$, Mat\'ern kernel \eqref{eq:Matern-1d} can be written in closed form. Let $q=\nu+1/2$, then the Mat\'erm-$\nu$ kernel with half-integer $\nu$ is the product of an exponential and a polynomial of order
\[k(x,x')\propto \exp({-{\omega}{\vert x-x'\vert}})\frac{q!}{2q!}\left(\sum_{l=0}^{q}\frac{(q+l)!}{l!(q-l)!}({2\omega}{\vert x-x'\vert})^{q-l}\right).\]
The key idea of our sparse formulations are based on two functions. The first one is that for any half-integer Mat\'ern-$\nu$ $k(\cdot,\cdot\vert\omega)$ kernel with scale parameter $\omega$ and any $2\nu+2$ points $x_1,\cdots,x_{2\nu+2}$ sorted in increasing order, there exists $2\nu+2$ coefficients   $a_1,\cdots,a_{2\nu+2}$ such that the following function
\[\phi_{(x_1,\cdots,x_{2\nu+2})}(\cdot)=\sum_{i=1}^{2\nu+2}a_ik(\cdot,x_i\vert\omega)\]
is non-zero only on the open interval $(x_1,x_{2\nu+2})$. The second one is that for the derivative of $k(\cdot,\cdot\vert\omega)$ and any $2\nu+2$ with respective to $\omega$ and $2\nu+4$ points $x_1,\cdots,x_{2\nu+4}$ sorted in increasing order, there exists $2\nu+4$ coefficients   $a_1,\cdots,a_{2\nu+4}$ such that the following function
\[\psi_{(x_1,\cdots,x_{2\nu+4})}(\cdot)=\sum_{i=1}^{2\nu+4}a_i\frac{\partial k(\cdot,x_i\vert\omega)}{\partial \omega}\]
is non-zero only on the open interval $(x_1,x_{2\nu+4})$. These two functions give rise to sparse factorization of covariance matrix and its derivative as product of banded matrix and inverse banded matrix. The first function is called Kernel packet (KP) and is derived in \cite{chen2022kernel}. The second one is the generalization of KP and, hence, we call it generalized KP.

\subsection{Covariance Matrix}
Using KP, we  can factorize any one-dimensional Mat\'ern covariance matrix $k_d(\BFX_d,\BFX_d)$ as product  of a banded matrix $\BFPhi_d$ and the inverse of a banded matrix  $\BFA_d$:
\begin{equation}
    \label{eq:banded_factorize}
    \BFP_d^Tk_d(\BFX_d,\BFX_d)\BFP_d=\BFA_d^{-1}\BFPhi_d
\end{equation}
where $\BFX_d$ is any one-dimensional point set, $\BFP_d$ is the permutation that sorts $\BFX_d$ in increasing order, and the band widths of both $\BFPhi_d$ and $\BFA_d$ are $\nu-1/2$ and $\nu+1/2$, respectively.

The basic idea of sparse factorization \eqref{eq:banded_factorize} relies on the following theorem regarding Mat\'erm kernel with half-integer smoothness parameter. The theorem summarizes central, right, and left KPs in \cite{chen2022kernel}.
\begin{figure}
    \centering
    \includegraphics[width=.4\linewidth]{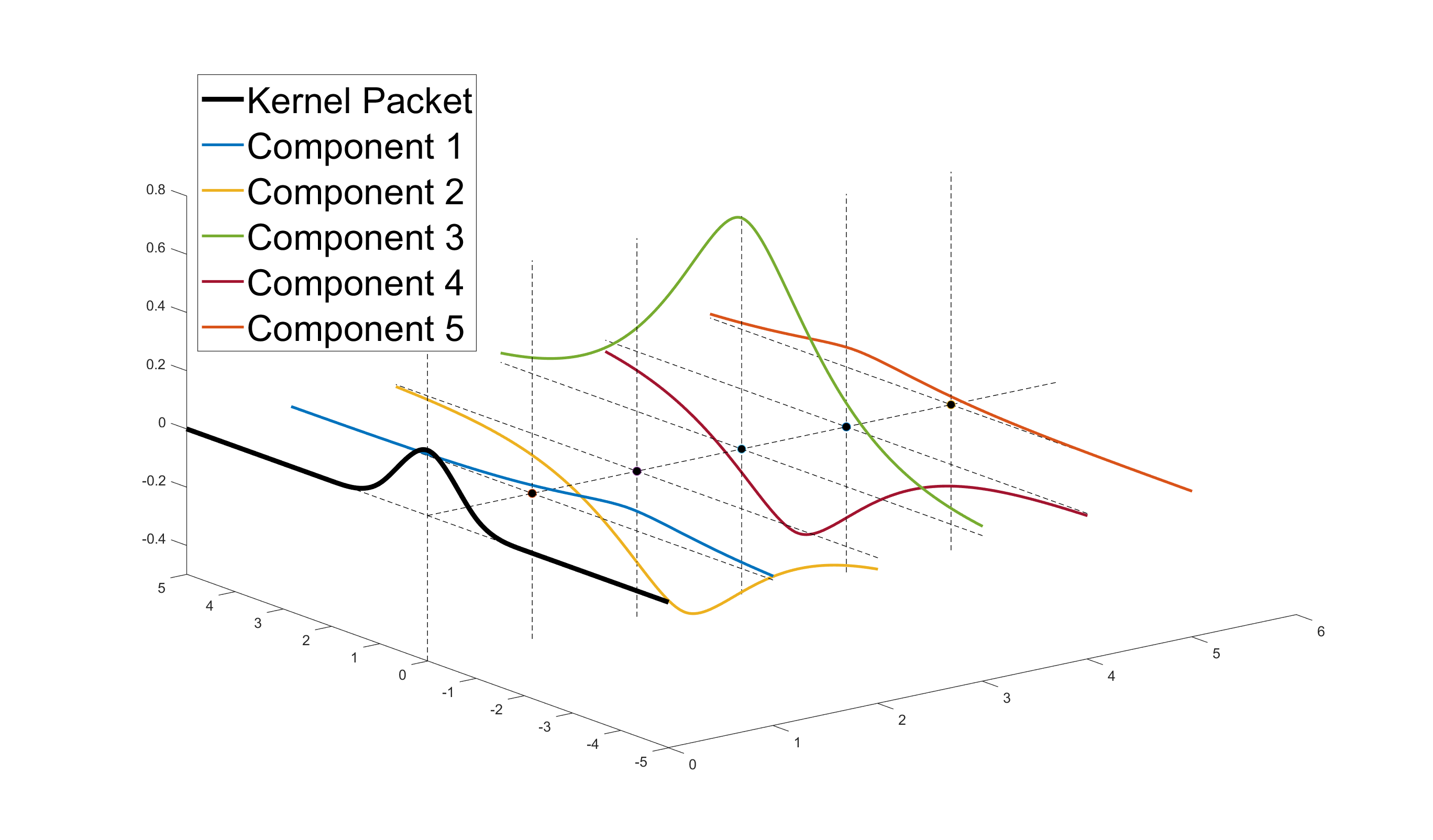}
    \includegraphics[width=.35\linewidth]{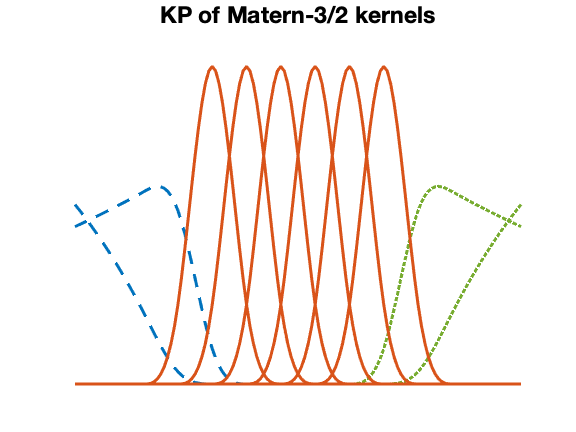}
    \caption{Left: the addition of five Mat\'ern-$\frac{3}{2}$ kernels $a_j k(\cdot,x_j)$ (colored lines, without compact supports) leads to 
    a KP (black line, with a compact support); Right: converting 10 Mat\'ern-$\frac{3}{2}$ kernel functions $\{k(\cdot,x_i)\}_{i=1}^{10}$ to 10 KPs, where each KP is non-zeron on at most three points in $\{x_i\}_{i=1}^{10}$.}
    \label{fig:KP}
\end{figure}
\begin{theorem}[\cite{chen2022kernel}]
\label{thm:KP}
Let $k$ be a Mat\'ern-$\nu$ kernel with half-integer smoothness parameter $\nu$, i.e., $\nu=\frac{1}{2}, \frac{3}{2},\cdots$. Let $(x_1<x_2<\cdots<x_p)$ be any sorted one-dimensional points.
 \begin{enumerate}
     \item  If $p=2\nu+2$, let $(a_1,\cdots,a_p)$ be the solution of the following system of equations:
    \begin{equation}\label{eq:KP}
        \sum_{i=1}^{p} a_i x_i^l \exp\{\delta c x_i\}=0,
    \end{equation}
with $l=0,\ldots,(p-1)/2$, $c=2\nu\omega^2/(2\pi)^2$, and $\delta=\pm 1$. Then function:
$\phi_{(x_1,\cdots,x_p)}=\sum_{i=1}^pa_ik(\cdot,x_i)$
is non-zero only on interval $(x_1,x_p)$;
    \item If  $\nu+\frac{3}{2}\leq p<2\nu+2$, let $(a_1,\cdots,a_p)$ be the solution of the following system of equations
    \begin{equation}\label{eq:KP_one_side}
    \sum_{i=1}^{p} a_ix_i^{l} \exp\{hcx_i\}=0,\quad \sum_{i=1}^{p} a_i x_i^{r} \exp\{-hcx_i\}=0,
\end{equation}
where $l=0,\ldots,(2\nu-1)/2$, and the second term comprises auxiliary equations with $r=0,\ldots,p-\nu-5/2$ ( if $p-\nu-5/2<0$, skip the right side of \eqref{eq:KP_one_side} ). If $h=1$, then function:
$\phi_{(x_1,\cdots,x_p)}=\sum_{i=1}^pa_ik(\cdot,x_i)$
is non-zero only on interval $(-\infty,x_p)$; If $h=-1$, then function:
$\phi_{(x_1,\cdots,x_p)}=\sum_{i=1}^pa_ik(\cdot,x_i)$
is non-zero only on interval $(x_1,\infty)$.
 \end{enumerate}
\end{theorem}
Theorem \ref{thm:KP} shows that for any $p$ sorted points $\{x_i\}_{i=1}^p$, there exists linear combination such that $\phi_{(x_1,\cdots,x_p)}=\sum_{i=1}^pa_ik(\cdot,x_i)$ is non-zero only on $(x_1,x_p)$. Remind that the associated one-dimensional GP covariance matrix $k_d(\BFX_d,\BFX_d)$ can be viewed as the values of $n$ kernel functions $\{k(\cdot,x_i)\}_{i=1}^n$ on $\{x_i\}_{i=1}^n$.  We can convert these $n$ kernel functions  to $n$ KPs and get the Gram matrix $\BFPhi$ on  $\{x_i\}_{i=1}^n$. Because each KP is  non-zero only at $p$ points in $\{x_i\}_{i=1}^n$, $\BFPhi$ must be a banded matrix. Algorithm \ref{alg:banded_factorization} shows the factorization \eqref{eq:banded_factorize} explicitly. 
\begin{algorithm}
\caption{Computing banded matrices $\BFA$ and $\BFPhi$ such that $\BFP^T\BFK\BFP=\BFA^{-1}\BFPhi$}\label{alg:banded_factorization}
\hspace*{\algorithmicindent} \textbf{Input}: one-dimension Mat\'ern-$\nu$ covariance matrix $\BFK$,  scattered points $\{x_{i}\}_{i=1}^n$ \\
 \hspace*{\algorithmicindent} \textbf{Output}: banded matrices $\BFA$ and $\BFPhi$, and permutation matrix $\BFP$ 
\begin{algorithmic}
\Ensure $\nu$ is a half-integer, $n \geq 2\nu+2$
\State Initialize $\BFA,\BFPhi \leftarrow \BFzero\in\Real^{n\times n}$
\State search permutation $\BFP$ to sort $\{x_{i}\}_{i=1}^n$ in increasing order

\For{$i= 1$ to $\nu+\frac{1}{2}$}
\State Compute $\{a_l\}_{l=1}^{i+\nu+1/2}$ associated to $\{x_l\}_{l=1}^{i+\nu+1/2}$ via \eqref{eq:KP_one_side} with $h=1$
 \State $[\BFA]_{1:i+\nu+1/2,i} \leftarrow (a_1,\cdots,a_{i+\nu+\frac{1}{2}})$\;
 %$[\BFPhi]_{1:i+\nu-1/2,i} \leftarrow \sum_{l=1}^{i+\nu+1/2}a_l[k(x_{1},x_{l}),\cdots,  k(x_{i+\nu-1/2},x_{l})]$\;
\EndFor
\For{$i= \nu+3/2$ to $n-\nu-\frac{1}{2}$}
\State Compute $\{a_l\}_{l=1}^{2\nu+2}$ associated to $\{x_l\}_{l=i-\nu-1/2}^{i+\nu+1/2}$ via \eqref{eq:KP} 
\State $[\BFA]_{i-\nu-1/2:i+\nu+1/2,i} \leftarrow (a_1,\cdots,a_{2\nu+2})$\;
 %$[\BFPhi]_{i-\nu+1/2:i+\nu-1/2,i} \leftarrow \sum_{l=1}^{2\nu+2}a_l[k(x_{i-\nu+1/2},x_{i-\nu-3/2+l}),\cdots,  k(x_{i+\nu-1/2},x_{i-\nu-3/2+l})]$\;
\EndFor
\For{$i= n-\nu+\frac{1}{2}$ to $n$}
\State
Compute $\{a_l\}_{l=1}^{n-i+\nu+3/2}$ associated to $\{x_l\}_{l=i-\nu-1/2}^{n}$  via \eqref{eq:KP_one_side} with $h=-1$
\State $[\BFA]_{i-\nu-1/2:n,i} \leftarrow (a_1,\cdots,a_{n-i+\nu+3/2})$\;
 %$[\BFPhi]_{i-\nu+1/2:n,i} \leftarrow \sum_{l=1}^{n-i+\nu+3/2}a_l[k(x_{i-\nu+1/2},x_{l}),\cdots,  k(x_{n},x_{l})]$\;
\EndFor
\State $\BFPhi=\BFA\BFP^T\BFK\BFP$
\end{algorithmic}
\end{algorithm}

We can analyze the time and space complexity of Algorithm \ref{alg:banded_factorization}. Firstly, sorting $n$ points in increasing order requires $\CalO(n\log n)$ time complexity. Secondly, the total $n$ iterations requires $\CalO(n)$ time complexity, as each iteration of the algorithm involves solving a $p\times p$ system of equations, which has a time complexity of $\CalO(1)$. We can also see that the matrices $\BFA$ is of band widths $\nu+1/2$ since at most $2\nu+2$ entries on the $i$-th row of $\BFA$ are flipped to non-zero in the $i$-th iteration. At last, the matrix $\BFPhi$ is a $\nu-1/2$-banded matrix for the $i$-th row of $\Phi$ equals to the value of a KP on $\{x_i\}_{i=1}^n$, which has at most $2\nu$ non-zero entries. Therefore, the time and space complexity for computing $\BFPhi=\BFA\BFK$ are both $\CalO(n)$. To summarize, the total time and space complexity of Algorithm \ref{alg:banded_factorization} are $\CalO(n\log n)$ and $\CalO(n)$, respectively.

\subsection{Derivative of Covariance Matrix}
In this subsection, we generalized the theory of KP in \cite{chen2022kernel} to show that for any Mat\'ern kernel with half-integer smoothness parameter $\nu$ and any $p=2\nu+4$ sorted points $\{x_i\}_{i=1}^p$, there exists linear combination such that
\[\psi_{(x_1,\cdots,x_p)}(x)=\sum_{i=1}^pb_i{\partial_\omega k(\omega \vert x-x_i\vert)}\]
is non-zero only on interval $(x_1,x_p)$. As a result, $\partial_\omega k_d(\omega\vert \BFX_d-\BFX_d\vert)$, the derivative of any one-dimensional Mat\'ern covariance matrix with respective to the scale hyperparameter $\omega$, can also be factorized as product of a banded patrix $\BFPsi_d$ and the inverse of a banded matrix $\BFB_d$:
\begin{equation}
    \label{eq:banded_factorize_gradient}
    \BFP_d^T\big[\partial_\omega k_d(\omega\vert \BFX_d-\BFX_d\vert)\big]\BFP_d=\BFB_d^{-1}\BFPsi_d
\end{equation}
where $\BFX_d$ is any one-dimensional point set, $\BFP_d$ is the permutation that sorts $\BFX_d$ in increasing order, and the band widths of both $\BFPsi_d$ and $\BFB_d$ are $\nu+1/2$ and $\nu+3/2$, respectively.
\begin{figure}
    \centering
    \includegraphics[width=.8\linewidth]{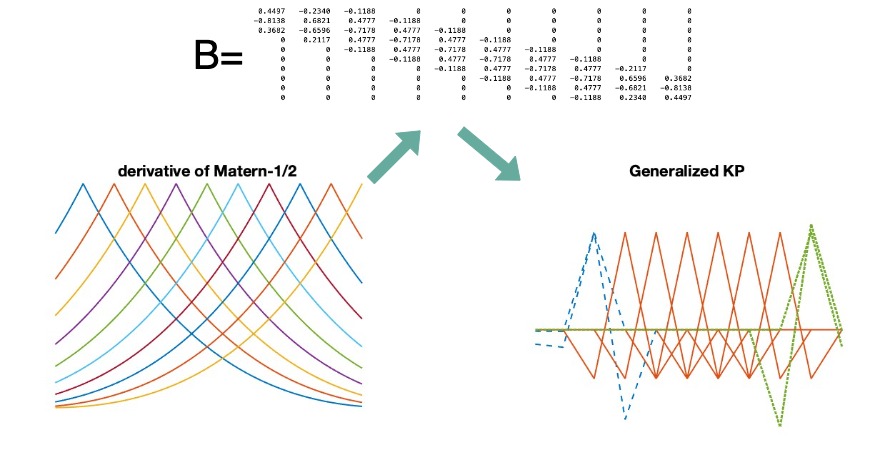}
    \caption{derivative of kernel functions: $\{{\partial_\omega k(\cdot,i/10)}\}_{i=1}^{10}$ can be converted to compactly support generalized KPs via a banded matrix $\BFA$}
    \label{fig:general_KP}
\end{figure}

Figure \ref{fig:general_KP} illustrates how to convert  $\partial_\omega k_d(\omega\vert \BFX_d-\BFX_d\vert)$, the derivative of Mat\'ern-$\frac{1}{2}$ matrix, with $\omega=1$ and $\BFX_d=\{.1,\cdots,1\}$ to the  banded matrix $\BFPsi_d$ via a specific banded matrix $\BFB$. The $(i,j)$-entry of $\partial_\omega k_d(\omega\vert \BFX_d-\BFX_d\vert)$ is equal to $\partial_\omega k_d(\omega\vert x_i-x_j\vert)=-\vert x_i-x_j\vert exp\{-\vert x_i-x_j\vert\}$ and, hence, the derivative matrix is dense.  On the other hand, the $(i,j)$-entry of $\BFPsi_d$ is equal to the value of generalized KP at $x_j$: $\psi_{(x_{i-2}, x_{i-1},x_{i},x_{i+1},x_{i+2})}(x_j)$. Therefore, $[\BFPsi]_{i,j}$ is equal to $0$ if $\vert i-j\vert\geq 2$ because in this case $x_j$ is out of the support of generalize KP $\psi_{(x_{i-2}, x_{i-1},x_{i},x_{i+1},x_{i+2})}$.

The construction of generalized KPs for the derivative of the Mat\'ern kernel is very similar, except we treat the derivative of a Mat\'ern-$\nu$ kernel as a Mat\'ern-$(\nu+1)$ kernel and use Theorem \ref{thm:KP} to compute the coefficients. The coefficients for constructing the KPs of the  Mat\'ern-$(\nu+1)$ kernel are also the coefficients for constructing the generalized KPs of the derivative of the Mat\'ern-$\nu$ kernel. Formal theories are left in Appendix. We directly show Algorithm \ref{alg:banded_factorization_gradient} for construction the sparse matrix factorization \eqref{eq:banded_factorize_gradient}. The time and space complexity of  Algorithm \ref{alg:banded_factorization_gradient} are exactly the same as Algorithm \ref{alg:banded_factorization}, which is $\CalO(n\log n)$. We use the following theorem to summarize Algorithm \ref{alg:banded_factorization_gradient}.

\begin{theorem}\label{thm:banded_factor_gradient}
    Let $\BFPsi$, $\BFB$ be the outputs of Algorithm \ref{alg:banded_factorization_gradient}. Then $\BFPsi$ is a $\nu+\frac{1}{2}$-banded matrix and $\BFB$ is a $\nu+\frac{3}{2}$-banded matrix. Moreover, for any set of scattered point $\{x_i\}_{i=1}^n$, $\BFB$ is invertable.
\end{theorem}

\begin{algorithm}
\caption{Computing banded matrices $\BFB$ and $\BFPsi$ such that $\partial_{\omega}\BFK_{\omega}=\BFB^{-1}\BFPhi$}\label{alg:banded_factorization_gradient}
\hspace*{\algorithmicindent} \textbf{Input}: derivative of Mat\'ern-$\nu$  covariance matrix $\partial_\omega\BFK_\omega$,  scattered points $\{x_{i}\}_{i=1}^n$ \\
 \hspace*{\algorithmicindent} \textbf{Output}: banded matrices $\BFB$ and $\BFPsi$, and permutation matrix $\BFP$ 
\begin{algorithmic}
\Ensure $\nu$ is a half-integer, $n \geq 2\nu+4$
\State Initialize $\BFB,\BFPsi \leftarrow \BFzero\in\Real^{n\times n}$
\State run Algorithm \ref{alg:banded_factorization} for Mat\'ern-$(\nu+1)$ covariance matrix on scatered points $\{x_i\}_{i=1}^n$
\State $\BFP$ is the output $\BFP$ of Algorithm \ref{alg:banded_factorization}
\State $\BFB\leftarrow\BFA$ where $\BFA$ is the output of Algorithm \ref{alg:banded_factorization}
\State $\BFPhi=\BFB\BFP^T\partial_\omega \BFK_\omega\BFP$
\end{algorithmic}
\end{algorithm}

In the following content, we will see that computing the gradient $\partial_\omega l$, we only need the banded matrices $\{\BFPsi_d\}_{d=1}^D$ and $\{\BFB^{-1}_d\}_{d=1}^D$ associated to one-dimensional covariance matrices $\{\BFK_d\}$. Therefore, Theorem \ref{thm:banded_factor_gradient} guarantees that all the computations and banded solvers applied are feasible.

\section{Fast Computation}
As described in \cite{hastie2009elements,gilboa2013scaling},  the backfitting algorithm is a widely used approach for fitting additive models in high-dimensional spaces. In this algorithm, the additive model is constructed by fitting individual univariate nonparametric regression models to each variable in the $D$-dimensional input space. The backfitting algorithm iteratively refines these models by alternately updating the fit for each variable while holding the others fixed. Each iteration of the backfitting algorithm involves solving a set of univariate regression problems, which can be done efficiently with methods like kernel smoothing or least squares regression. For polynomial smoothers or Gauss-Markov models, each iteration requires $\CalO(n)$ time complexity. However, current backfittig algorithms for additive GPs are only for prediction, i.e., computing the posterior mean in \eqref{eq:GP_posterior}. To the best of our knowledge, direct application of backfitting algorithm  on computing the posterior variance in \eqref{eq:GP_posterior}, the log-likelihood function $l$ in \eqref{eq:add_likelihood} and its gradient in \eqref{eq:add_likelihood_gradient} is still unknown.

Our algorithms, which are essentially based on iterative method,  can compute the posterior mean, posterior variance, the log-likelihood $l$, and the gradient of $l$ efficiently. Before presenting our algorithm, we first directly substitute sparse factorization \eqref{eq:banded_factorize} into \eqref{eq:posterior_block} and \eqref{eq:add_likelihood}, and  substitute sparse factorization \eqref{eq:banded_factorize_gradient} into\eqref{eq:add_likelihood_gradient} to rewrite them in the following forms:
\begin{align}
    &\mu_n(\BFx^*)=\boldsymbol{1}^T\BFphi^T({\BFx^*})\BFPhi^{-T}\BFP^T[\BFP\BFPhi^{-1}\BFA\BFP^T+\frac{1}{\sigma_y^2}\BFS\BFS^T]^{-1}\BFS(\frac{1}{\sigma_y^2}\rmI_n)\BFY\label{eq:posterior_block_sparse_mean} \\
    &s_n(\BFx^*)=\sum_{d=1}^Dk_d(x^*_d,\BFx^*_d)-\sum_{d=1}^D\BFphi^T_d(x^*_d)\BFPhi_d^{-T}\BFA_d^{-1}\BFphi_d(x^*_d)\nonumber\\
    &\quad\quad+\boldsymbol{1}^T\BFphi^{-T}({\BFx^*})\BFPhi^{-T}\BFP^T[\BFP\BFPhi^{-1}\BFA\BFP^T+\frac{1}{\sigma_y^2}\BFS\BFS^T]^{-1}\BFP\BFPhi^{-1}\BFphi({\BFx^*})\boldsymbol{1}\label{eq:posterior_block_sparse_var}\\
    &l(\BFomega,\sigma_y)\propto\frac{\BFY}{D}^T\BFS^T\BFP\BFA_{\BFthete}^T\BFPhi_{\BFthete}^{-T}\BFP^T[\BFP\BFPhi_{\BFthete}^{-1}\BFA_{\BFthete}\BFP^T+\frac{1}{\sigma_y^2}\BFS\BFS^T]^{-1}\BFP\BFPhi_{\BFthete}^{-1}\BFA_{\BFthete}\BFP^T\BFS\frac{\BFY}{D}\nonumber\\
    &\quad\quad-\frac{\BFY}{D}^T\BFS^T\BFP\BFA_{\BFthete}^T\BFPhi_{\BFthete}^{-T}\BFP^T\BFS\frac{\BFY}{D}-\log\vert\BFPhi_{\BFthete}^{-1}\BFA_{\BFthete}+\frac{1}{\sigma_y^2}\BFS\BFS^T\vert \nonumber\\
    &\quad\quad-\log\vert\BFPhi_{\BFthete}\vert+\log\vert\BFA_{\BFthete}\vert-2n\log\sigma_y\label{eq:posterior_block_sparse_l}\\
    &\frac{\partial l}{\partial \omega_d}\propto\BFY^TR\BFB_d^{-1}\BFPsi_dR\BFY-\text{Trace}\left[R\BFB_d^{-1}\BFPsi_d\right]\label{eq:posterior_block_sparse_gradient}
\end{align}
where
\begin{align*}
    &\BFPhi=\begin{bmatrix}
\BFPhi_1 & &\\
 &\ddots &\\
&  & \BFPhi_D
\end{bmatrix},\quad \BFA=\begin{bmatrix}
\BFA_1 &  &\\
&\ddots &\\
& & \BFA_D
\end{bmatrix},\quad\BFP=\begin{bmatrix}
\BFP_1 & &\\
 &\ddots &\\
 & & \BFP_D
\end{bmatrix}\in\Real^{Dn\times Dn}\\
& \BFphi(\BFx^*)=\BFA\boldsymbol{\gamma}_{\BFx^*}=\begin{bmatrix}
\BFphi_1(x^*_1) & & \\
 &\ddots &\\
 & & \BFphi_D(x^*_D)
\end{bmatrix}\in\Real^{Dn\times D},\\
& R=\frac{1}{\sigma_y^2}\rmI_n-\frac{1}{\sigma_y^2}\BFS^T[\BFP\BFPhi^{-1}\BFA\BFP^T+\frac{1}{\sigma_y^2}\BFS\BFS^T]^{-1}\BFS\frac{1}{\sigma_y^2},
\end{align*}
$\{\BFP_d\}$ are the permutation matrices, 
$\{\BFPhi_d\}$ and $\{\BFA_d\}$ are the banded matrices in factorization \eqref{eq:banded_factorize} for one-dimensional GP covariance matrix $k_d(\BFX_d,\BFX_d)$, $\{\BFPsi_d\}$ and $\{\BFB_d\}$ are the banded matrices in factorization \eqref{eq:banded_factorize_gradient}, $\BFphi_d(x_d^*)=\BFA_dk_d(\BFX_d,x^*_d)$ are values of KPs at the $d$-th dimension of input point $\BFx^*$, and $\BFPhi_{\BFthete}$ and $\BFA_{\BFthete}$ are the $\BFPhi$ and $\BFA$ induced by hyperparameters $\BFthete$, respectively. The factorizations in \eqref{eq:posterior_block_sparse_mean},\eqref{eq:posterior_block_sparse_var},\eqref{eq:posterior_block_sparse_l}, and \eqref{eq:posterior_block_sparse_gradient} play key roles in our algorithms. In the following content, we first introduce the training part of our our algorithm and then the prediction part.

 \subsection{Training}\label{sec:training}
In the training part, we propose a fast algorithm for matrix inverse and a fast algorithm for matrix determinant. We first introduce the algorithm for matrix inverse then the algorithm for  matrix determinant.
\subsubsection{Matrix Inverse}
In the training part, all computations involving matrix inverse are among one of the following three operations 
\begin{enumerate}
    \item Computing vector $\BFb_{\BFv}=\boldsymbol{T}^{-1}\BFv$ for vector $\BFv$ where $\boldsymbol{T}$ is banded matrix $\BFPhi$ or $\BFB_d$; \label{enu:operation_0}
    \item compute the inverse matrix $\BFM:= [\BFP\BFPhi^{-1}\BFA\BFP^T+\frac{1}{\sigma_y^2}\BFS\BFS^T]^{-1}$; \label{enu:operation_1}
    \item compute  vector $\BFb_{\BFv}:=[\BFP\BFPhi^{-1}\BFA\BFP^T+\frac{1}{\sigma_y^2}\BFS\BFS^T]^{-1}\BFv$ for any vector $\BFv$ \label{enu:operation_2} ;
    \item compute the $(\nu+\frac{1}{2})$-band of matrix $\BFPhi_d^{-T}\BFA_d^{-1}$ for $d=1,\cdots,D$ \label{enu:operation_3}.
\end{enumerate}
Operation \ref{enu:operation_0} can be done in $\CalO(n)$ time by applying  banded matrix solver. For example, the algorithm based on the LU decomposition in \cite{davis2006direct} can be applied to solve the equation  $\boldsymbol{T}^{-1}\boldsymbol{b}=\BFv$ in $\CalO(n)$ time, where $\BFb$ is the unknown.  MATLAB provides convenient and efficient builtin functions, such as \texttt{mldivide} or \texttt{decomposition}, to solve sparse banded linear system in this form. 

Operation \ref{enu:operation_1} and \ref{enu:operation_3} are not necessary in many situations. However, if we want to  reduce the computation time of $s_n(\BFx^*)$ from $\CalO(n\log n)$to $\CalO(\log n)$ for randomly selected $\BFx^*$ after training, full knowledge of $\BFM$ and the  $(\nu+\frac{1}{2})$-band of matrix $\BFPhi_d^{-T}\BFA_d^{-1}$ must be given. For predetermined predictive point $\BFx^*$,  operation \ref{enu:operation_1} and \ref{enu:operation_3} can be skipped and only operation \ref{enu:operation_2} 
 and \ref{enu:operation_3} will involve in the whole computation process, which requires $\CalO(n\log n)$ time in total. The reason for operation \ref{enu:operation_3} is that only the  $(\nu+\frac{1}{2})$-band of matrix $\BFPhi_d^{-T}\BFA_d^{-1}$ is required in the later prediction part.

\begin{algorithm}
\caption{Computing vector $\tilde{\BFv}:=[\BFP\BFPhi^{-1}\BFA\BFP^T+\frac{1}{\sigma_y^2}\BFS\BFS^T]^{-1}\BFv$}\label{alg:inverse_v}
\hspace*{\algorithmicindent} \textbf{Input}: permutation matrices $\{\BFP_d\}_{d=1}^D$, banded matrices $\{\BFPhi_{d}\}_{d=1}^D$ and $\{\BFA_d\}_{d=1}^D$, vector $\BFv$, observational noise variance $\sigma_y^2$\\
 \hspace*{\algorithmicindent} \textbf{Output}:  vector $\tilde{\BFv}$
\begin{algorithmic}
\State Initialize $\tilde{\BFv}_{d}^{(0)}\leftarrow\BFzero\in\Real^{n}$, $d=1,\cdots,D$
\For{$t= 1$ to $T$}
\State \begin{equation}
    \label{eq:Gauss_Seidel_v}
    \tilde{\BFv}_{d}^{(t+1)}=\BFP_d^T\BFPhi_d[\sigma_y^2\BFA_d+\BFPhi_d]^{-1}\BFP_d\left(\frac{1}{\sigma_y^2}\BFv_d-\sum_{d'<d}\tilde{\BFv}_{d'}^{(t+1)}-\sum_{d'>d}\tilde{\BFv}_{d'}^{(t)}\right)
\end{equation}
\State where $\BFv_d$ denotes the $(dn-n+1)$-th entry to the $dn$-th entry of $\boldsymbol{v}$
 %$[\BFPhi]_{1:i+\nu-1/2,i} \leftarrow \sum_{l=1}^{i+\nu+1/2}a_l[k(x_{1},x_{l}),\cdots,  k(x_{i+\nu-1/2},x_{l})]$\;
\EndFor
\State return $\tilde{\BFv}=[\tilde{\BFv}_{d}^{(T)}]_{d}$
\end{algorithmic}
\end{algorithm}

Fast computations of operation \ref{enu:operation_1} and \ref{enu:operation_2}  rely on Algorithm \ref{alg:inverse_v}. Algorithm \ref{alg:inverse_v} is based on the Gauss-Seidel method \citep{davis2006direct} for computing the vector $\tilde{\BFv}=[\BFK+\sigma_y^{-2}\BFS\BFS^T]^{-1}\BFv$. What we have done is to decompose the following iterative step into block matrix operations:
\begin{align}
    &\quad\begin{bmatrix}
    \sigma_y^2 \BFK_1^{-1} + \rmI_n & & & &\\
 \rmI_n&\sigma_y^2\ \BFK_2^{-1} + \rmI_n& & &\\
\vdots& & &\ddots &\\
 \rmI_n& \cdots& & \rmI_n& \sigma_y^2 \BFK_1^{-1} +\rmI_n
\end{bmatrix}\tilde{\BFv}^{(t+1)}\nonumber\\
&=\sigma_y^{-2}\BFv-
\begin{bmatrix}
\BFzero & \rmI_n & \rmI_n &\cdots& \rmI_n\\
\BFzero& \BFzero& \rmI_n &\cdots & \rmI_n\\
& &\ddots & &\\
& & & &\BFzero
\end{bmatrix}\tilde{\BFv}^{(t)}.\label{eq:Gauss_Seidel_direct}
\end{align}
Compared with direct Gauss-Seidel \eqref{eq:Gauss_Seidel_direct}, the improvement in the iterative step \eqref{eq:Gauss_Seidel_v} is that it can be computed in $\CalO(n)$ time complexity since $[\sigma_y^2\BFA_d+\BFPhi_d]$ is a banded matrix and we can apply the banded matrix solver in operation \ref{enu:operation_0} to compute \eqref{eq:Gauss_Seidel_direct} .  

In practice,the number of iterations required in Algorithm \ref{alg:inverse_v} is far less than the number of data and usually set in the order of $\CalO(\log n)$ or even $\CalO(1)$ \citep{saatcci2012scalable}. As a result, the over time complexity of Algorithm \ref{alg:inverse_v} is $\CalO(n)$. %Similarly,  we can see that $\tilde{\BFM}$ is multiplication of $\BFM$ by inverse banded matrices and permutation matrices. We can compute $\tilde{\BFM}$ in $\CalO(n^2)$ time by using these fast algorithms for banded matrix.

Operation \ref{enu:operation_2} can be computed directly by applying Algorithm \ref{alg:inverse_v} on associated vector. Inverse matrix $\BFM$ in operation \ref{enu:operation_1} can be computed by applying Algorithm \ref{alg:inverse_v} on vectors $\BFe_1,\cdots,\BFe_n$ where $\BFe_i$ is a zero vector with $1$ on its $i$-th entry. The $i$-th column of $\BFM$ is exactly the output of Algorithm \ref{alg:inverse_v} with input $\BFe_i$, i.e., 
\[\BFM_{:,i}=[\BFP\BFPhi^{-1}\BFA\BFP^T+\frac{1}{\sigma_y^2}\BFS\BFS^T]^{-1}\BFe_i.\]
Because we need to apply Algorithm \ref{alg:inverse_v} $n$ times for computing matrix $\BFM$, the computational time complexity for operation \ref{enu:operation_1} is $\CalO(n^2 )$.
%Algorithm \ref{alg:inverse_v} is a slight modification of Algorithm \ref{alg:inverse_M}. Different from Algorithm \ref{alg:inverse_M}, which aims to solve a dense matrix, the objective of Algorithm \eqref{alg:inverse_v} is to solve for a vector.  After solving vector $\tilde{\BFv}$ using Algorithm \ref{alg:inverse_v}, vector $\BFb_{\BFv}=\BFPhi^{-T}\BFP^T\tilde{\BFv}$ can be computed in $\CalO(n)$ time using banded matrix solver. To summarize, both of them are based on the Gauss-Seidel method. Convergence of these two algorithms is improved significantly by the fact that we are jointly updating blocks of $n$ variables at a time. Consequently, in practice, the number of iterations required is far less than updating the unknowns one by one. 

\begin{algorithm}
\caption{Computing the $(\nu+\frac{1}{2})$-band of $\BFPhi^{-T}_d\BFA_d^{-1}$}\label{alg:band_PhiA}
\hspace*{\algorithmicindent} \textbf{Input}: banded matrices $\BFPhi_{d}$ and $\BFA_d$ \\
 \hspace*{\algorithmicindent} \textbf{Output}:  $[\BFPhi^{-T}_d\BFA_d^{-1}]_{i,j}$ for $|i-j|\leq \nu+\frac{1}{2}$
\begin{algorithmic}
\State Define matrix blocks $\BFH_i^-,\BFH_i,\BFH_i^+$ of $[h_{i,j}]:=\BFA_d\BFPhi_d
^T$ as 
\begin{align}
    \BFH_i^-=&\begin{bmatrix}
 h_{s_i,s_i-2\nu}  &\cdots& h_{s_i,s_i-1}\\
   &\ddots &  \vdots \\
    & &h_{s_{i+1}-1,s_{i}-1}
\end{bmatrix},\nonumber\\
\BFH_i=&\begin{bmatrix}
 h_{s_i,s_i}  &\cdots& h_{s_i,s_{i+1}-1}\\
  \vdots &\ddots &  \vdots \\
  h_{s_{i+1}-1,s_{i}} &\cdots &h_{s_{i+1}-1,s_{i+1}-1}
\end{bmatrix},\nonumber\\
\BFH_i^+=&\begin{bmatrix}
 h_{s_i,s_{i+1}}  & & \\
  \vdots &\ddots &   \\
  h_{s_{i+1}-1,s_{i+1}} &\cdots &h_{s_{i+1}-1,s_{i+2}-1}
\end{bmatrix}\label{eq:tridiagonal_block}
\end{align}
where $i=1,\cdots,I$, $I=\lceil \frac{n}{2\nu}\rceil$, $s_i=(i-1)2\nu+1$, and $s_{I+1}-1=\min\{n,2\nu I \}$ \State \Comment{{{$\BFA_d\BFPhi_d
^T$ is a $2\nu$-banded matrix, and $\BFH_1^{-}$ and $\BFH_{I}^+$ are null}}}
%$[\BFPhi^{-1}_d\BFA_d^{-1}]_{i,j}$ for $i=1,\cdots,2\nu$, $j=1,\cdots,4\nu$
\State Define matrix blocks $\BFM_i^-,\BFM_i,\BFM_i^+$ of $\BFPhi^{-T}_d\BFA_d^{-1}$ corresponding to the same entry indices of $\BFH_i^-,\BFH_i,\BFH_i^+$
\State Solve $\BFM_1$, $\BFM_1^+$
\For{$j= 2$ to $I$}
\State $\BFM_j^-=\BFM_{j-1}^+$ \Comment{{$\BFA_d\BFPhi_d^T=\BFA_d\BFK_d\BFA^T_d$ is a symmetric matrix}}
%Denote $\BFM=[\BFPhi^{-1}_d\BFA_d^{-1}]_{(i\nu-\nu+1):i\nu,(i\nu-\nu+1):i\nu}, \BFM_-=[\BFPhi^{-1}_d\BFA_d^{-1}]_{(i\nu-\nu+1):i\nu,(i\nu-2\nu+1):(i-1)\nu}, \BFM_+=[\BFPhi^{-1}_d\BFA_d^{-1}]_{(i\nu-\nu+1):i\nu,(i\nu-\nu+1):i\nu}$\;
 %$[\BFPhi]_{1:i+\nu-1/2,i} \leftarrow \sum_{l=1}^{i+\nu+1/2}a_l[k(x_{1},x_{l}),\cdots,  k(x_{i+\nu-1/2},x_{l})]$\;
 \State Solve auxiliary matrix $\BFM_j^{--}$: $$\BFH_{j-1}^{-}\BFM_{j-2}+\BFH_{j-1}\BFM_{j-1}^-+\BFH_{j-1}^+\BFM_j^{--}=\BFzero$$ \Comment{skip for $j=2$}
 \State Solve $\BFM_j: \BFM_j^{--}\BFH_{j-1}^-+\BFM_j^{-}\BFH_{j-1}+\BFM_j\BFH_{j-1}^+=\BFzero$
 \State Solve $\BFM_{j}^+: \BFM_j^{-}\BFH_{j}^-+\BFM_j\BFH_{j}+\BFM_j^{+}\BFH_{j}^+=\rmI_{2\nu}$ \Comment{skip for $j=I$}
\EndFor
\State return $\BFM_j^-,\BFM_j,\BFM_j^+, j=1,\cdots I$
\end{algorithmic}
\end{algorithm}

\begin{figure}
    \centering
    \includegraphics[width=.3\linewidth]{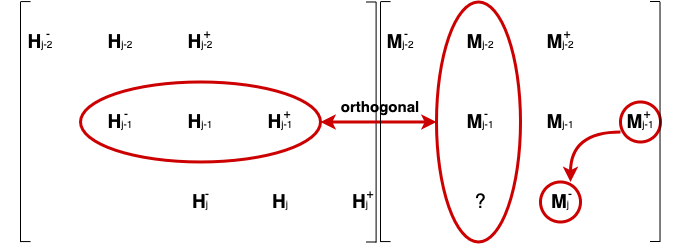}
    \includegraphics[width=.3\linewidth]{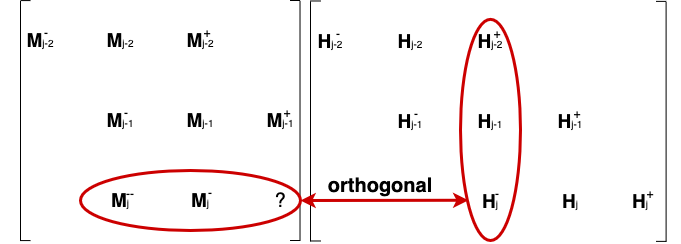}
    \includegraphics[width=.3\linewidth]{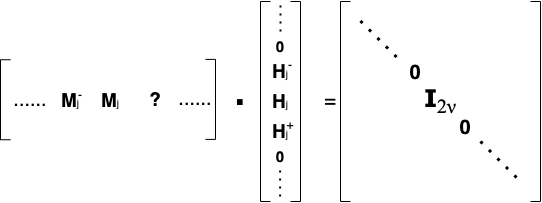}
    \caption{The time complexity of calculating any block matrix is $\CalO(\nu^3)$, provided that it can be placed in a consecutive row/column with two other known block matrices. This is because $\BFH$ is a block-tridiagonal matrix. When working on the $j$-th column, we can get $\BFM_{j}^-=\BFM_{j-1}^+$ directly by symmetry and solve an auxiliary matrix $\BFM_j^{--}$ by putting $[\BFM_{j-2};\BFM_{j-1}^-;\BFM_j^{--}]$ in a consecutive column (left); then we can first put $[\BFM_j^{--}, \BFM_j^- ,\BFM_j]$ to solve $\BFM_j$ (middle), and then put $[\BFM_j^{-}, \BFM_j ,\BFM_j^+]$ to solve $\BFM_j^+$ (right). }
    \label{fig:banded_inverse}
\end{figure}

Algorithm \ref{alg:band_PhiA} is designed to perform operation \ref{enu:operation_3} in $\CalO(\nu^2n)$ time. The main concept behind Algorithm \ref{alg:band_PhiA} is that the multiplication of a $(\nu + 1/2)$-banded matrix with a $(\nu - 1/2)$-banded matrix results in a $2\nu$-banded matrix, which can be partitioned into a block-tridiagonal matrix $\BFH = \text{diag}[\BFH^-_j, \BFH_j, \BFH^+_j]$, where each block is a $2\nu$-by-$2\nu$ matrix. Since we only require the $(\nu + 1/2)$-band of $\BFPhi^{-T}_d\BFA_d^{-1}$, we can utilize the block-tridiagonal property of $\BFH$. This means that the multiplication of any row/column of  $\BFPhi^{-T}_d\BFA_d^{-1}$ by any column/row of $\BFH$ only involves three consecutive $2\nu$-by-$2\nu$ block matrices from $\BFPhi^{-T}_d\BFA_d^{-1}$. The process of computing the band of $\BFPhi^{-T}_d\BFA_d^{-1}$ is illustrated in Figure \ref{fig:banded_inverse}. Solving a $2\nu$-by-$2\nu$ matrix equation has a time complexity of $O(\nu^3)$, and since we only need to solve $O(n/\nu)$ of these matrix equations, the total time complexity of Algorithm \ref{alg:band_PhiA} is $O(\nu^2n)$.

\subsubsection{Matrix Determinant and Trace }
Remind that for learning the hyperparameter, we must also compute the log-likelihood function \eqref{eq:posterior_block_sparse_l} and its gradient \eqref{eq:posterior_block_sparse_gradient}, which involve computing matrix inverse, matrix determinants and matrix trace. All Matrix inversions in \eqref{eq:posterior_block_sparse_l} and \eqref{eq:posterior_block_sparse_gradient} can be computed by using Algorithm \eqref{alg:inverse_v} and banded matrix solver. We also need to compute the following terms:
\[\log\vert\BFPhi_{\BFthete}^{-1}\BFA_{\BFthete}+\frac{1}{\sigma_y^2}\BFS\BFS^T\vert,
    \quad\log\vert\BFPhi_{\BFthete}\vert,\quad \log\vert\BFA_{\BFthete}\vert, \quad \text{Trace}\left[R\BFB_d^{-1}\BFPsi_d\right].\]
The log determinant terms $\log\vert\BFPhi_{\BFthete}\vert$ and $\log\vert\BFA_{\BFthete}\vert$ can also be computed in $\CalO(\nu^2n)$ time 
because both $\BFA_{\BFthete}$ and $\BFPhi_{\BFthete}$ are $\nu+1/2$-banded matrix and $\nu-1/2$-banded matrix, respectively, and their determinants can be computed in $\CalO(\nu^2n)$ time by sequential methods \citep{kamgnia2014some}. 

For $\log\vert\BFPhi_{\BFthete}^{-1}\BFA_{\BFthete}+\frac{1}{\sigma_y^2}\BFS\BFS^T\vert$, we first estimate its largest eigenvalue, and then we utilize the Taylor expansion of matrix log determinant to design an iterative method that resembles Algorithm \ref{alg:inverse_v}, which enables us to estimate its value.

\begin{algorithm}
\caption{[Power Method] Computing the largest eigenvalue of $\BFPhi_{\BFthete}^{-1}\BFA_{\BFthete}+\frac{1}{\sigma_y^2}\BFS\BFS^T$ }\label{alg:power_method}
\hspace*{\algorithmicindent} \textbf{Input}: matrix $\BFPhi_{\BFthete}^{-1}\BFA_{\BFthete}+\frac{1}{\sigma_y^2}\BFS\BFS^T$\\
 \hspace*{\algorithmicindent} \textbf{Output}:  largest eigenvalue of $\BFPhi_{\BFthete}^{-1}\BFA_{\BFthete}+\frac{1}{\sigma_y^2}\BFS\BFS^T$
\begin{algorithmic}
\State Initialize $ \lambda_{\max}\leftarrow\BFzero\in\Real^{Dn}$
\For{$q= 1$ to $Q$}
\State Initialize $ {\BFv}^{(0)}$: each entry of $ {\BFv}^{(0)}$ is uniformly distributed on $\{-1,1\}$
\For{$s= 1$ to $S$}
\State \begin{equation}
    \label{eq:power_method}
     {\BFv}^{(s)}=[\BFPhi_{\BFthete}^{-1}\BFA_{\BFthete}+\frac{1}{\sigma_y^2}\BFS\BFS^T] {\BFv}^{(s-1)}
\end{equation}
 %$[\BFPhi]_{1:i+\nu-1/2,i} \leftarrow \sum_{l=1}^{i+\nu+1/2}a_l[k(x_{1},x_{l}),\cdots,  k(x_{i+\nu-1/2},x_{l})]$\;
\EndFor
\State $\lambda = [\BFv^{(S)}]^T[\BFPhi_{\BFthete}^{-1}\BFA_{\BFthete}+\frac{1}{\sigma_y^2}\BFS\BFS^T] {\BFv}^{(S)}/\|\BFv^{(S)}\|^2$
\State \textbf{If} $\lambda>\lambda_{max}$: $\lambda_{\max}\leftarrow \lambda$
\EndFor
\State return $\lambda_{max}$
\end{algorithmic}
\end{algorithm}

To estimate the largest eigenvalue of $\BFPhi_{\BFthete}^{-1}\BFA_{\BFthete}+\frac{1}{\sigma_y^2}\BFS\BFS^T$, we can simply use the power method \cite{mises1929praktische} as shown in Algorithm \ref{alg:power_method}. Because both $\BFPhi_{\BFthete}$ and $\BFA_{\BFthete}$ are banded matrices, \eqref{eq:power_method} in each iteration can be computed in $\CalO(n)$ time using LU decomposition as discussed before. The number of iterations required in Algorithm \ref{alg:power_method}  is independent of the data size $n$ for power method is essentially a Monte Carlo method.

After we estimate the largest eigenvalue of $\BFPhi_{\BFthete}^{-1}\BFA_{\BFthete}+\frac{1}{\sigma_y^2}\BFS\BFS^T$, we can normalize it to a positive definite matrix with eigenvalue less than 1 and then apply the following Taylor expansion of log determinant to the normalized matrix:
\begin{equation}
    \label{eq:taylor_log_det}
    \log \vert M\vert=-\sum_{s=1}^\infty \frac{1}{s}\text{trace}\left((\rmI-M)^s\right)
\end{equation}
where $M$ is any positive definite matrix with all eigenvalues less than $1$. The basic idea of computing the log determinant of $\BFPhi_{\BFthete}^{-1}\BFA_{\BFthete}+\frac{1}{\sigma_y^2}\BFS\BFS^T$ numerically is to use a truncation of \eqref{eq:taylor_log_det}.  The key ingredient for fast computation of \eqref{eq:taylor_log_det} is efficient computation of the trace $\text{trace}\left((\rmI-M)^s\right)$. Algorithm \eqref{alg:trace_rand}  in \cite{avron2011randomized} is a randomized algorithm for computing the trace of any symmetric positive definite (SPD) matrix. It has been proved in \cite{avron2011randomized} that the total number of iterations $Q$ required for a certain level of precision is independent of matrix size $n$. So time complexity of Algorithm \ref{alg:trace_rand} only depends on the time complexity for computing matrix multiplication \eqref{eq:trace_estimate}.

\begin{algorithm}
\caption{[\cite{avron2011randomized}] Computing  the trace of a matrix $\BFM$ }\label{alg:trace_rand}
\hspace*{\algorithmicindent} \textbf{Input}: any SPD matrix $\BFM\in\Real^{n \times n}$\\
 \hspace*{\algorithmicindent} \textbf{Output}:   trace of $\BFM$
\begin{algorithmic}
\State $\gamma\leftarrow 0$

\For{$q= 1$ to $Q$}
\State Initialize $\BFv_q\sim \CalN(0,\rmI_{n})$
\State \begin{equation}
    \label{eq:trace_rand}
    \gamma\leftarrow \BFv_q^T\BFM\BFv_q+\gamma
\end{equation}
 %$[\BFPhi]_{1:i+\nu-1/2,i} \leftarrow \sum_{l=1}^{i+\nu+1/2}a_l[k(x_{1},x_{l}),\cdots,  k(x_{i+\nu-1/2},x_{l})]$\;
 %\State $\gamma\leftarrow \frac{1}{s}\BFv_0^T\BFv_s+\gamma$
\EndFor
\State $\gamma\leftarrow \frac{\gamma}{Q}$
\State return $\gamma$
\end{algorithmic}
\end{algorithm}

Algorithm \ref{alg:log_det} combines Algorithm \ref{alg:power_method} and Algorithm \ref{alg:trace_rand} to numerically compute the log determinant of a matrix. The inner iteration number $S$ is in fact the truncated order of the Taylor expansion \eqref{eq:taylor_log_det}:
\begin{equation}\label{eq:taylor_log_det_trunc}
    \log\vert \frac{1}{\lambda_{max}} \left(\BFPhi_{\BFthete}^{-1}\BFA_{\BFthete}+\frac{1}{\sigma_y^2}\BFS\BFS^T\right)\vert\approx -\sum_{s=1}^S\frac{1}{s}\text{trace}\left(\frac{\BFPhi_{\BFthete}^{-1}\BFA_{\BFthete}+\frac{1}{\sigma_y^2}\BFS\BFS^T}{\lambda_{max}}\right)
\end{equation}
and the outer iteration number $Q$ is the number of samples for estimating the trace of a matrix. It has been proved in \cite{boutsidis2017randomized} that the truncation \eqref{eq:taylor_log_det_trunc} converges to the true value exponential fast in truncation point $S$. Therefore, only $S=\CalO(\log n)$ is required for the inner iteration in Algorithm \ref{alg:log_det}. Remind that $Q$ is independent of data size $n$ and \eqref{eq:trace_estimate} can be computed in $\CalO(n)$ time by banded matrix solver, Algorithm \ref{alg:log_det} require $\CalO(n\log n)$ time.

\begin{algorithm}
\caption{Computing   $\log\vert \BFPhi_{\BFthete}^{-1}\BFA_{\BFthete}+\frac{1}{\sigma_y^2}\BFS\BFS^T\vert$ }\label{alg:log_det}
\hspace*{\algorithmicindent} \textbf{Input}: matrix $\BFPhi_{\BFthete}^{-1}\BFA_{\BFthete}+\frac{1}{\sigma_y^2}\BFS\BFS^T$\\
 \hspace*{\algorithmicindent} \textbf{Output}:   $\log\vert \BFPhi_{\BFthete}^{-1}\BFA_{\BFthete}+\frac{1}{\sigma_y^2}\BFS\BFS^T\vert$
\begin{algorithmic}
\State use Algorithm \ref{alg:power_method} to estimate $\lambda_{max}$, the largest eigenvalue of  $ \BFPhi_{\BFthete}^{-1}\BFA_{\BFthete}+\frac{1}{\sigma_y^2}\BFS\BFS^T$
\State $\BFM\leftarrow \rmI_{Dn}-\frac{1}{\lambda_{max}} \left(\BFPhi_{\BFthete}^{-1}\BFA_{\BFthete}+\frac{1}{\sigma_y^2}\BFS\BFS^T\right)$
\State $\gamma\leftarrow 0$
\For{$q= 1$ to $Q$}
\State Initialize $\BFv_0\sim \CalN(0,\rmI_{Dn})$
\For{$s= 1$ to $S$}
\State \begin{equation}
    \label{eq:trace_estimate}
    \BFv_s=\BFM\BFv_{s-1}
\end{equation}
 %$[\BFPhi]_{1:i+\nu-1/2,i} \leftarrow \sum_{l=1}^{i+\nu+1/2}a_l[k(x_{1},x_{l}),\cdots,  k(x_{i+\nu-1/2},x_{l})]$\;
 \State $\gamma\leftarrow \frac{1}{s}\BFv_0^T\BFv_s+\gamma$
\EndFor
\EndFor
\State $\gamma\leftarrow nD\log \lambda_{max}-\frac{\gamma}{Q}$
\State return $\gamma$
\end{algorithmic}
\end{algorithm}

For computing the trace term $\text{Trace}\left[R\BFB_d^{-1}\BFPsi_d\right]$, we can directly use Algorithm \ref{alg:trace_rand}. For any $\BFv_q$ in \eqref{eq:trace_rand}, we have
\begin{align}
    \BFv_q^TR\BFB_d^{-1}\BFPsi_d\BFv_q=\frac{\BFv_q^T\BFB_d^{-1}\BFPsi_d\BFv_q}{\sigma_y^2}-\frac{\BFv^T}{\sigma_y^2}\BFS^T[\BFP\BFPhi^{-1}\BFA\BFP^T+\frac{1}{\sigma_y^2}\BFS\BFS^T]^{-1}\BFS\frac{\BFB_d^{-1}\BFPsi_d\BFv_q}{\sigma_y^2}\label{eq:trace_term}
\end{align}
The first term of \eqref{eq:trace_term} can be computed in $\CalO(n)$ using banded matrix solver. For the second term, we can first use bander matrix solver to get $\BFB_d^{-1}\BFPsi_d\BFv_q$ and then use Algorithm \ref{alg:inverse_v}, which take $\CalO(n)\log n$ time in total. Also, the number of iterations $Q$ in Algorithm \ref{alg:trace_rand} is independent of $n$ and, hence, the time complexity for computing the trace is $\CalO(n\log n)$

To summarize, the log determinant terms $\log \vert \BFphi_{\BFthete}\vert$ and $\log \vert\BFA_{\BFthete}\vert$ can both be computed in $\CalO(\nu^2n)$ time using LU decomposition algorithm, while $\log\vert \BFPhi_{\BFthete}^{-1}\BFA_{\BFthete}+\frac{1}{\sigma_y^2}\BFS\BFS^T\vert$ can be computed in $\CalO(n\log n)$ time using Algorithm \ref{alg:log_det} and $\text{Trace}\left[R\BFB_d^{-1}\BFPsi_d\right]$ can be computed in $\CalO(n\log n)$ using Algoritm \ref{alg:trace_rand}. Therefore, the total time complexity for computing all the determinant terms in the log-likelihood function \eqref{eq:posterior_block_sparse_l} and its gradient is $\CalO(n\log n)$.

\subsection{Prediction}\label{sec:prediction}
Because the support of KP $[\BFphi_d]_i$ is $(x_{i-\nu-1/2,d},x_{i+\nu+1/2,d})$ for any $i=1,\cdots,n$ and $d=1,\cdots,D$, so, for any predictive point $\BFx^*$ and dimension $d$, there must be at most $2\nu+1$ non-zero entries on $\BFphi_d(x^*_d)$ and these non-zero entries are also consecutive. This fact is the essential idea for fast computation of prediction. For a given $x^*_d$, searching the $2\nu+1$ consecutive non-zero entries on $\BFphi_d(x^*_d)$ is equivalent to searching the sorted data point $x_{i,d}$ such that $x_{i,d}<x^*_d<x_{i+1,d}$, which requires $\CalO(\log n)$ time complexity only.

For fixed hyperhyparameter $\BFthete$ and predetermined predictive point $\BFx^*$, we can skip operation \ref{enu:operation_1} and do operation \ref{enu:operation_2} and 3 only in the training. In this case, we can first use Algorithm \ref{alg:inverse_v} and fast banded matrix solver to compute vectors $\BFb_{\BFv_1}:=\BFPhi^{-T}\BFP^T[\BFP\BFPhi^{-1}\BFA\BFP^T+\frac{1}{\sigma_y^2}\BFS\BFS^T]^{-1}\BFS(\frac{1}{\sigma_y^2}\rmI_n)\BFY\nonumber$ and $\BFb_{\BFv_2}:=\BFPhi^{-T}\BFP^T[\BFP\BFPhi^{-1}\BFA\BFP^T+\frac{1}{\sigma_y^2}\BFS\BFS^T]^{-1}\BFP\BFPhi^{-1}\BFphi({\BFx^*})$ in $\CalO(n\log n)$ time. For  $\mu_n(\BFx^*)$ in \eqref{eq:posterior_block_sparse_mean}, because $\mu_n(\BFx^*)$ equals sparse vector multiplication $\boldsymbol{1}^T\BFphi^T({\BFx^*})\BFb_{\BFv_1}$, the prediction part only required $\CalO(1)$ time complexity. For $s_n(\BFx^*)$ in \eqref{eq:posterior_block_sparse_var}, the third term equals $\boldsymbol{1}^T\BFphi^T({\BFx^*})\BFb_{\BFv_2}$, which is also sparse matrix multiplication and can be computed in $\CalO(1)$ time. For the third term, we first use Algorithm \ref{alg:band_PhiA}
 to get the $\nu+\frac{1}{2}$-band of $\BFPhi_d^{-T}\BFA_d^{-1}$ for $d=1,\cdots,D$. This operation requires $\CalO(n)$ time. Suppose the $2\nu+1$  consecutive non-zeros entries on $\BFphi(\BFx^*)$ are with indices $\{i_d^*,i^*_d+1,\cdots,i^*+2\nu\}_{d=1}^D$, then the second term of $s_n(\BFx^*)$ is
 \begin{align}
     &\quad \sum_{d=1}^D\BFphi_d^T(x^*_d)\BFPhi_d^{-T}\BFA^{-1}_d\BFphi_d(x^*_d)
     \nonumber\\&=\sum_{d=1}^D\sum_{j=0}^{2\nu}\sum_{l=0}^{2\nu}[\BFphi_d(x^*_d)]_{i_d^*+j}[\BFphi_d(x^*_d)]_{i_d^*+l}[\BFPhi_d^{-T}\BFA^{-1}_d]_{i_d^*+j,i_d^*+l}.\label{eq:sparse_sum_phiPAphi}
 \end{align}
Obviously, there are at $D4\nu^2$ additions in the \eqref{eq:sparse_sum_phiPAphi}, and, as a result, computation of the second term is also $\CalO(1)$.

To summarize, the total time complexity for the case that hyperparameters and predictive point are fixed before training, the overall time complexity is $\CalO(n\log n)$ because operation \ref{enu:operation_2} and 3 in training only requires $\CalO(n\log n)$ time and, except for searching the non-zero entries on $\BFphi(\BFx^*)$, which requires $\CalO(\log n)$ time, every other step in prediction part only requires $\CalO(1)$ time.  

However, in many application, hyperparameters and predictive point cannot be known before training. For example, in Bayesian optimization, we need to estimate the hyperparameter for the model and then search an optimal point based on the posterior. In this case, we must compute the matrix $\BFM$ and $\tilde{\BFM}$ in operation \ref{enu:operation_1} of the training part. Although the time complexity for computing $\BFM$ and $\tilde{\BFM}$ is increased to $\CalO(n^2)$ compared with the previous case, time complexity for prediction part remains unchanged. Notice that vector  $\BFb_{\BFv_1}:=\BFPhi^{-T}\BFP^T[\BFP\BFPhi^{-1}\BFA\BFP^T+\frac{1}{\sigma_y^2}\BFS\BFS^T]^{-1}\BFS(\frac{1}{\sigma_y^2}\rmI_n)\BFY\nonumber$ is independent of predictive point $\BFx^*$ so computation of $\mu_n(\BFx^*)$ remains unchanged. Also, the sparsity in \eqref{eq:sparse_sum_phiPAphi} remains unchanged. The only difference is that we now need to compute the term $\boldsymbol{1}^T\BFphi^T({\BFx^*})\tilde{\BFM}\BFphi({\BFx^*})\boldsymbol{1}$ with out using Algorithm \ref{alg:inverse_v} for otherwise every prediction require $\CalO(n)$ time. Similary to \eqref{eq:sparse_sum_phiPAphi}, we can see that the computation of $\BFphi^T({\BFx^*})\tilde{\BFM}\BFphi({\BFx^*})$ only involves $4D^2\nu^2$ addition as follows:
\[\sum_{d,d'=1}^D\BFphi_d^T({x^*_d})\tilde{\BFM}_{d,d'}\BFphi_{d'}({x^*_{d'}})\]
where $[\tilde{M}_{d,d'}]_{d,d'}$ is the partition of $\tilde{\BFM}$ into $D \times D$ block matrices and for any $(d,d')$
\begin{align}
     &\quad\BFphi_d^T(x^*_d)\tilde{M}_{d,d'}\BFphi_{d'}(x^*_{d'})\nonumber\\&=\sum_{j=0}^{2\nu}\sum_{l=0}^{2\nu}[\BFphi_d(x^*_d)]_{i^*_d+j}[\BFphi_{d'}(x^*_{d'})]_{i_{d'}^*+l}[\tilde{M}_{d,d'}]_{i_d^*+j,i_{d'}^*+l}. \label{eq:sparse_sum_phiMphi}
 \end{align}

 To summarize, the total time complexity for unknown hyperparameters and predictive point is $\CalO(n^2)$ because operation \ref{enu:operation_1} and computing $\partial l/\partial\BFthete$ involve multiplications of dense matrices, which at least requires $\CalO(n^2)$ time. Nonetheless, every step in prediction part remains unchanged, which only requires $\CalO(1)$ time. 

 \subsection{Summary}
We have proposed different algorithms for computing different terms in the posterior mean \eqref{eq:posterior_block_sparse_mean}, posterior variance \eqref{eq:posterior_block}, the log-likelihood \eqref{eq:posterior_block_sparse_l} and its gradient \eqref{eq:posterior_block_sparse_gradient}. To summarize the efficient computation of additive GPs, we first list all the terms we need to compute and use a table to briefly review their efficient computations.  
\begin{align*}
    &\mu_n(\BFx^*)=\boldsymbol{1}^T\BFphi^T({\BFx^*})\underbrace{\BFPhi^{-T}\BFP^T[\BFP\BFPhi^{-1}\BFA\BFP^T+\frac{1}{\sigma_y^2}\BFS\BFS^T]^{-1}\BFS(\frac{1}{\sigma_y^2}\rmI_n)\BFY}_{\BFb_{\BFY}} \\
    &s_n(\BFx^*)=\sum_{d=1}^Dk_d(x^*_d,\BFx^*_d)-\sum_{d=1}^D\BFphi^T_d(x^*_d)\underbrace{\BFPhi_d^{-T}\BFA_d^{-1}}_{\nu+1/2\ \text{band}}\BFphi_d(x^*_d)\nonumber\\
    &\quad\quad+\boldsymbol{1}^T\BFphi^{-T}({\BFx^*})\underbrace{\BFPhi^{-T}\BFP^T[\BFP\BFPhi^{-1}\BFA\BFP^T+\frac{1}{\sigma_y^2}\BFS\BFS^T]^{-1}\BFP\BFPhi^{-1}}_{\BFM}\BFphi({\BFx^*})\boldsymbol{1} \\
    &l(\BFthete)\propto\underbrace{\frac{\BFY}{D}^T\BFS^T\BFP\BFA_{\BFthete}^T\BFPhi_{\BFthete}^{-T}\BFP^T[\BFP\BFPhi_{\BFthete}^{-1}\BFA_{\BFthete}\BFP^T+\frac{1}{\sigma_y^2}\BFS\BFS^T]^{-1}\BFP\BFPhi_{\BFthete}^{-1}\BFA_{\BFthete}\BFP^T\BFS\frac{\BFY}{D}}_{\text{quad}-\mathcal{A}}\nonumber\\
    &\quad\quad-\underbrace{\frac{\BFY}{D}^T\BFS^T\BFP\BFA_{\BFthete}^T\BFPhi_{\BFthete}^{-T}\BFP^T\BFS\frac{\BFY}{D}}_{\text{quad}-\mathcal{B}}-\underbrace{\log\vert\BFPhi_{\BFthete}^{-1}\BFA_{\BFthete}-\frac{1}{\sigma_y^2}\BFS\BFS^T\vert}_{\text{log\ det}-\mathcal{A}} \nonumber\\
    &\quad\quad-\underbrace{\log\vert\BFPhi_{\BFthete}\vert}_{\text{log\ det}-\mathcal{B}}+\underbrace{\log\vert\BFA_{\BFthete}\vert}_{\text{log\ det}-\mathcal{C}}-2n\log\sigma_y\\
    &\frac{\partial l}{\partial \BFthete_d}\propto\underbrace{\BFY^TR\BFB_d^{-1}\BFPsi_dR\BFY}_{\text{quad}-\mathcal{D}}-\underbrace{\text{Trace}\left[R\BFB_d^{-1}\BFPsi_d\right]}_{\text{Trace}}.
\end{align*}

\begin{table}[ht]
\caption{ Summary of computations for the posterior and log-likelihood }\label{tab:summary}
\centering
{
\begin{tabular}{| c| c|c| }
 \hline
 Term & Algorithm &Time Complexity\\
 \hline
$\BFb_{\BFy}$   &\specialcell{ Algorithm \ref{alg:inverse_v} and \\ LU decomposition    }& $\CalO(n\log n)$\\
\hline
$ \BFphi^T(\BFx^*)\BFb_{\BFy}$   & \specialcell{sparse vector\\ multiplication}   & \specialcell{$\CalO(1)$ if  $\BFb_{\BFy}$ and non-zero entries on $\BFphi(\BFx^*)$\\ are known; $\CalO(\log n)$ if  $\BFb_{\BFy}$ is known \\ but non-zero entries on $\BFphi(\BFx^*)$ are unknown}\\
\hline
\specialcell{$(\nu+1/2$)-band of \\ $\BFPhi^{-T}_d\BFA^{-1}_d$\\[.5em] } & Algorithm \ref{alg:band_PhiA} & $\CalO(\nu^2 n)$\\
\hline
\specialcell{$\BFphi_d^T(x^*_d)\BFPhi^{-T}_d$\\ $\cdot\BFA^{-1}_d\BFphi_d(x^*_d)$}&\specialcell{Sparse vector\\ multiplication or \\ LU decomposition}& \specialcell{$\CalO(1)$ if the $(\nu+1/2$)-band of  $\BFPhi^{-T}_d\BFA^{-1}_d$ \\and non-zero entries on $\BFphi(\BFx^*)$ are known;\\$\CalO(\log n)$ if the $(\nu+1/2$)-band of  $\BFPhi^{-T}_d\BFA^{-1}_d$ \\ are known but non-zero entries on $\BFphi(\BFx^*)$\\ are unknown; $\CalO(n\log n)$ if the $(\nu+1/2$)-\\band of  $\BFPhi^{-T}_d\BFA^{-1}_d$ are unknown }\\
\hline
$\BFM$& \specialcell{ Algorithm \ref{alg:inverse_v} and \\ LU decomposition    } &$\CalO(n^2)$\\
\hline
$\BFphi^T(\BFx^*)\BFM\BFphi(\BFx^*)$ &\specialcell{sparse vector\\ multiplication or \\Algorithm \ref{alg:inverse_v}}& \specialcell{$\CalO(1)$ if  $\BFM$ and non-zero entries on $\BFphi(\BFx^*)$\\ are known; $\CalO(\log n)$ if  $\BFM$ is known \\ but non-zero entries on $\BFphi(\BFx^*)$ is unknown;\\ $\CalO(n\log n)$ if $\BFM$ is unknown}\\
\hline
quad-$\mathcal{A},\mathcal{B}$& \specialcell{ Algorithm \ref{alg:inverse_v} and \\ LU decomposition    } & $\CalO(n\log n)$\\
\hline
log det-$\mathcal{A}$& \specialcell{ Algorithms \ref{alg:power_method}, \ref{alg:log_det}, and \\ LU decomposition    } & $\CalO(n\log n)$\\
\hline
log det-$\mathcal{B},\mathcal{C}$& \specialcell{sequential method\\ in \cite{kamgnia2014some} } & $\CalO(\nu^2n)$\\
\hline
%$R\BFY$&\specialcell{ Algorithm \ref{alg:inverse_v} and \\ LU decomposition    } & $\CalO(n\log n)$\\
%$R$ &\specialcell{ Algorithm \ref{alg:inverse_v}    } & $\CalO(n^2)$\\
%\hline
quad-$\mathcal{D}$ &\specialcell{Algorithm \ref{alg:inverse_v} and LU decomposition}& \specialcell{$\CalO(n\log n)$ }\\\hline
$\text{Trace}$ &\specialcell{Algorithm \ref{alg:inverse_v} and \ref{alg:trace_rand}  }& \specialcell{$\CalO(n\log n)$}\\\hline
\end{tabular}}
\end{table}

\section{Application to Bayesian Optimization}
 In this section, we use the sparsity nature of KP $\BFphi(\cdot)$ to design a fast algorithm for searching the next sampling point in Bayesian optimization, i.e., searching $\BFx=\arg\max_{\BFx}A(\BFx,\BFX,\BFY)$. In short, write $A:=A(\cdot,\BFX,\BFY)$. We use GP-UCB as an example to explicitly demonstrate our algorithm. Our algorithm can also be applied to general acquisition function, such as EI, as we will discuss later. Remind that the acquisition function of GP-UCB is
 \begin{equation}
     \label{eq:GP-UCB}
     A(\BFx)=\mu_n(\BFx)+\beta_n\sqrt{s_n(\BFx)}
 \end{equation}
 where, in our setting, $\mu_n$ is the posterior mean and $s_n$ is the posterior variance of an additive GP with Mat\'ern covariance. Updating the posterior in Algorithm \ref{alg:BayesOpt} can be done by the training procedure in Section \ref{sec:training}. We mainly introduce the gradient method for searching the maximizaer. 

 From Section \ref{sec:prediction}, \eqref{eq:GP-UCB} can be written in a sparse form for any predictive point $\BFx^*$:
 \begin{align}
A(\BFx^*)&=\underbrace{\sum_{d=1}^D\sum_{i=i^*}^{i*+2\nu}\BFb_{i,d}[\BFphi_d(x^*_d)]_i}_{\mu_n(\BFx^*)}\nonumber\\&\quad+\beta_n\underbrace{\sqrt{\sum_{d,d'=1}^D\sum_{i,j=i^*}^{i^*+2\nu}[\BFphi_d(x^*_d)]_i[\BFphi_{d'}(x^*_{d'})]_j m_{i,j,d,d'}}}_{\sqrt{s_n(\BFx^*)}}\label{eq:GP_UCB_sparse}
 \end{align}
 where $\BFb_{i,d}$ is the entry on vector $\BFPhi^{-T}\BFP^T[\BFP\BFPhi^{-1}\BFA\BFP^T+\frac{1}{\sigma_y^2}\BFS\BFS^T]^{-1}\BFS(\frac{1}{\sigma_y^2}\rmI_n)\BFY\nonumber$ corresponding to $[\BFphi_d(x^*_d)]_i$ on $\BFphi(\BFx^*)$ in vector multiplication and, similarly, $m_{i,j,d,d'}$ is the entry on matrix $\BFPhi^{-T}\BFA^{-1}+\BFPhi^{-T}\BFP^T[\BFP\BFPhi^{-1}\BFA\BFP^T+\frac{1}{\sigma_y^2}\BFS\BFS^T]^{-1}\BFP\BFPhi^{-1}$ corresponding to $[\BFphi_d(x^*_d)]_i$ on $\BFphi^T(\BFx^*)$ and $[\BFphi_{d'}(x^*_{d'})]_j$ on $\BFphi(\BFx^*)$ in the quadratic form. Remind that, for any $d$, there are only $\CalO(1)$ non-zero entries on KPs $\BFphi_d(x_d^*)$. As a result, computing the gradient of $A$ at $\BFx^*$ only requires $\CalO(1)$ operations
 \begin{align}
     \frac{\partial A}{\partial x^*_d}=& \frac{\partial \mu_n}{\partial x^*_d}+\beta_n \frac{\partial \sqrt{s_n}}{\partial x^*_d}\label{eq:GP_UCB_sparse_gradient}
 \end{align}
 where 
\begin{align*}
    &\frac{\partial \mu_n}{\partial x^*_d}=\sum_{i=i^*}^{i*+2\nu}\BFb_{i,d}\frac{\partial[\BFphi_d(x^*_d)]_i}{\partial x^*_d}\\
    &\frac{\partial \sqrt{s_n}}{\partial x^*_d}=\frac{1}{2\sqrt{s_n(\BFx^*)}}\bigg(2\sum_{d=1}^D\sum_{i,j=i^*}^{i^*+2\nu}\frac{\partial[\BFphi_d(x^*_d)]_i}{\partial x^*_d}[\BFphi_{d'}(x^*_{d'})]_j m_{i,j,d,d}\\
      &\quad\quad+\sum_{d\neq d'}^D\sum_{i,j=i^*}^{i^*+2\nu}\frac{\partial[\BFphi_d(x^*_d)]_i}{\partial x^*_d}[\BFphi_{d'}(x^*_{d'})]_j m_{i,j,d,d'}\bigg)
\end{align*}
and the gradient of a KP at $x_d^*$ is
 \[\frac{\partial [\BFphi_d(x^*_d)]_i}{\partial x_d^*}=\sum_{j=i^*-\nu-\frac{1}{2}}^{i^*+\nu+1/2}[\BFA_d]_{j,i}\frac{\partial k_d(x_j,x^*_d)}{\partial x^*_d}.\]
Recall that $s_n(\BFx^*)$ is summation of finite term in \eqref{eq:GP_UCB_sparse}, we can see that the number of additions in \eqref{eq:GP_UCB_sparse_gradient}  is independent of the total number of data $n$. The gradient of $\mu_n(\BFx^*)$ and $s_n(\BFx^*)$ can be written in the following more compact matrix forms:
\begin{align}
    &\nabla_{\BFx^*} \mu_n(\BFx^*)=\boldsymbol{g}_{\footnotesize \BFphi}^T({\BFx^*})\BFPhi^{-T}\BFP^T\BFM\BFS(\frac{1}{\sigma_y^2}\rmI_n)\BFY,\nonumber\\
    &\nabla_{\BFx^*} s_n(\BFx^*)=\boldsymbol{g}_{\footnotesize \BFphi}^T({\BFx^*})\left[-\BFPhi^{-T}\BFA^{-1}+2\BFPhi^{-T}\BFP^T\BFM\BFP\BFPhi^{-1}\right] \boldsymbol{g}_{\footnotesize \BFphi}({\BFx^*})\boldsymbol{1}, \label{eq:gradient_matrix}
\end{align}
where 
\begin{align*}
    &\boldsymbol{g}_{\footnotesize \BFphi}({\BFx^*})=\begin{bmatrix}
{\partial_{x^{*}_1}}\BFphi_1(x^*_1) & & &\\
&{\partial_{x^{*}_2}}\BFphi_2(x^*_2)& &\\
& &\ddots &\\
& & &{\partial_{x^{*}_D}} \BFphi_D(x^*_D)
\end{bmatrix}\in\Real^{Dn\times D},  \\
    &\BFM=[\BFP\BFPhi^{-1}\BFA\BFP^T+\frac{1}{\sigma_y^2}\BFS\BFS^T]^{-1}\in\Real^{Dn\times Dn},
\end{align*}
vector ${\partial_{x^{*}_d}}\BFphi_d(x^*_d)$ consists of derivative of $d$-dimensional KPs $\BFphi_d$ at $x^*_d$, and $\boldsymbol{1}=[1,\cdots,1]^T$ is a $D$-vector with all entries equal $1$. \eqref{eq:gradient_matrix} can be derived directly by applying matrix derivative laws on \eqref{eq:posterior_block_sparse_mean} and \eqref{eq:posterior_block_sparse_var}. Please refer to \cite{petersen2008matrix} for matrix calculus in details. So the gradient of $A$ in matrix form is
\[\nabla_{\BFx^*}A(\BFx^*)= \nabla_{\BFx^*} \mu_n(\BFx^*)+\frac{\beta_n}{2\sqrt{s_n(\BFx^*)}} \nabla_{\BFx^*} s_n(\BFx^*).\]

In general, an acquisition function $A(\BFx^*)$ is a composition function of the form $A(\BFx^*,\mu_n(\BFx^*),s_n(\BFx^*))$ because the Gaussian posterior can be fully determined by the posterior mean $\mu_n$ and posterior variance $s_n$. In this case, the gradient of $A$ is
\[\nabla_{\BFx^*}A=[\frac{\partial A}{\partial x^*_1}, \cdots,\frac{\partial A}{\partial x^*_D}]^T+\frac{\partial A}{\partial \mu_n}\nabla_{\BFx^*} \mu_n(\BFx^*)+\frac{\partial A}{\partial s_n}\nabla_{\BFx^*} s_n(\BFx^*).\]
Because the derivatives of $A$ with respective to $x^*_d$, $\mu_n$, and $s_n$ are all independent of data  size $n$, they all can be computed in $\CalO(1)$ time. The gradient $\nabla_{\BFx^*} \mu_n(\BFx^*)$ and $\nabla_{\BFx^*} s_n(\BFx^*)$ can also be computed in $\CalO(1)$ time as discussed previously. We can conclude that for any acquisition function $A$, its gradient with respective to $\BFx^*$ can be computed in $\CalO(1)$ time if our method is applied.

Indeed, when using gradient method for searching the maximizer of $A$, the time complexity of computing the posterior at each updated point $\BFx^{**}=\BFx^*+\delta\nabla_{\BFx^*} A(\BFx^*)$ can be further reduced to $\CalO(1)$ when the learning rate $\delta $ satisfies 
\[ \delta\|\nabla_{\BFx^*} A(\BFx^*)\|\leq {C}{\min_{i,j,d}\vert x_{i,d}-x_{j,d}\vert}\]
where $C$ is some constant independent of $n$. In this case, the updated point $\BFx^{**}$ is not a randomly selected new predictive point but a point near $\BFx^*$. Remind that the supports of KPs $\{[\BFphi_d]_i\}_{i=1}^n$ are consecutive, it is straightforward to derive that indices for non-zero entries on $\BFphi_d(x^{**}_d)$ are within the $C=\frac{\|\delta\nabla_{\BFx^*} A(\BFx^*,\BFX,\BFY)\|}{\min_{i,j,d}\vert x_{i,d}-x_{j,d}\vert}$-nearest neighbors of   indices for non-zero entries on $\BFphi_d(x^{*}_d)$. As a result, we can search the non-zero entries on $\BFphi_d(\BFx^{**}_d)$ by looking at the approximately  all the $C$-nearest neighbors of $\BFx^*$,   which requires $\CalO(1)$ time only. In generally, computing the value of acquisition $A$ at $\BFx^{**}$ requires $\CalO(\log n)$ time. This is because we need to search the $x_{i,d}$ that is closet to $x^{**}_d$ for all dimension $d$ as we have explained in section \ref{sec:prediction}.

\section{Numerial Experiments}

\begin{figure}
    \centering
    \includegraphics[width=.45\linewidth]{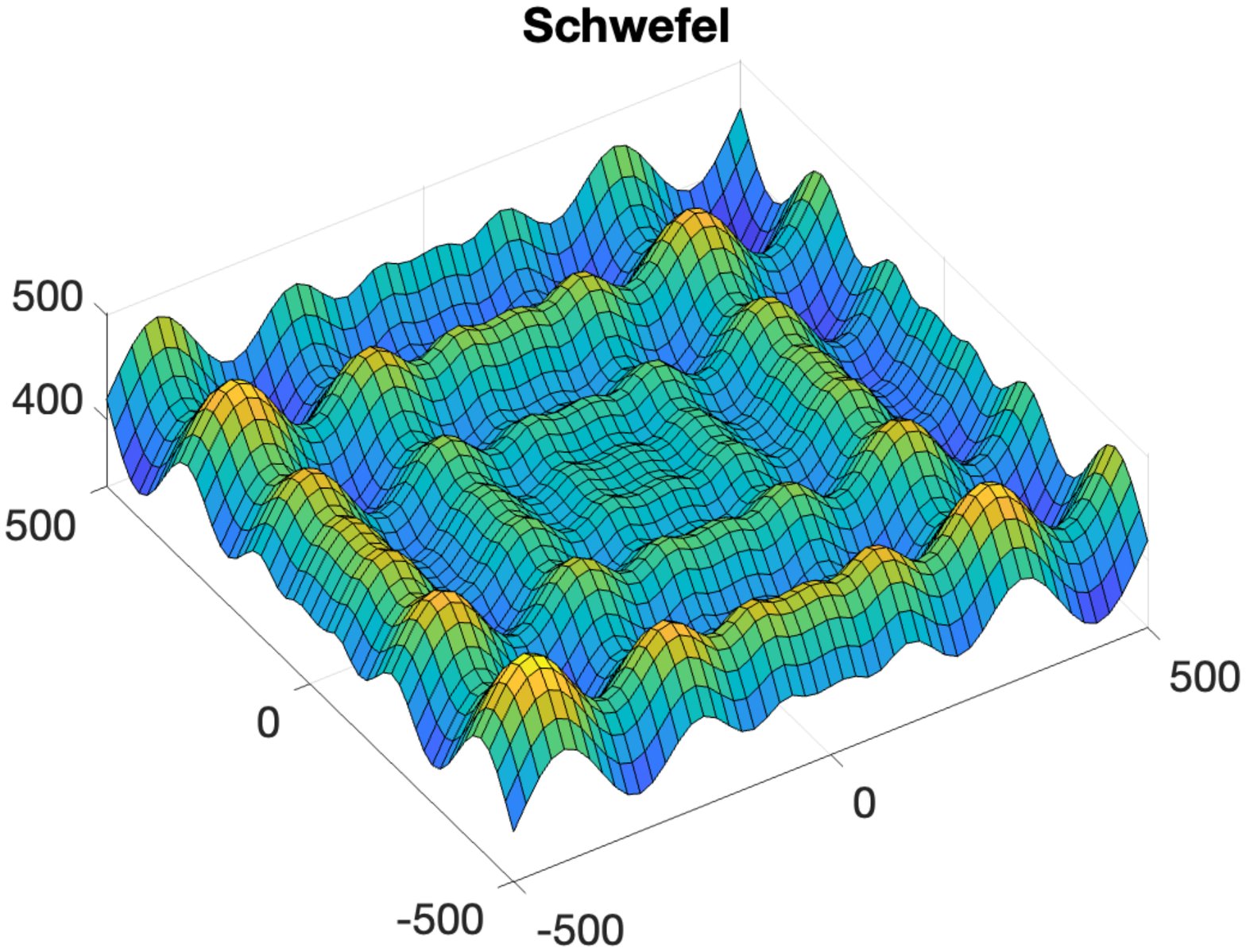}
    \includegraphics[width=.45\linewidth]{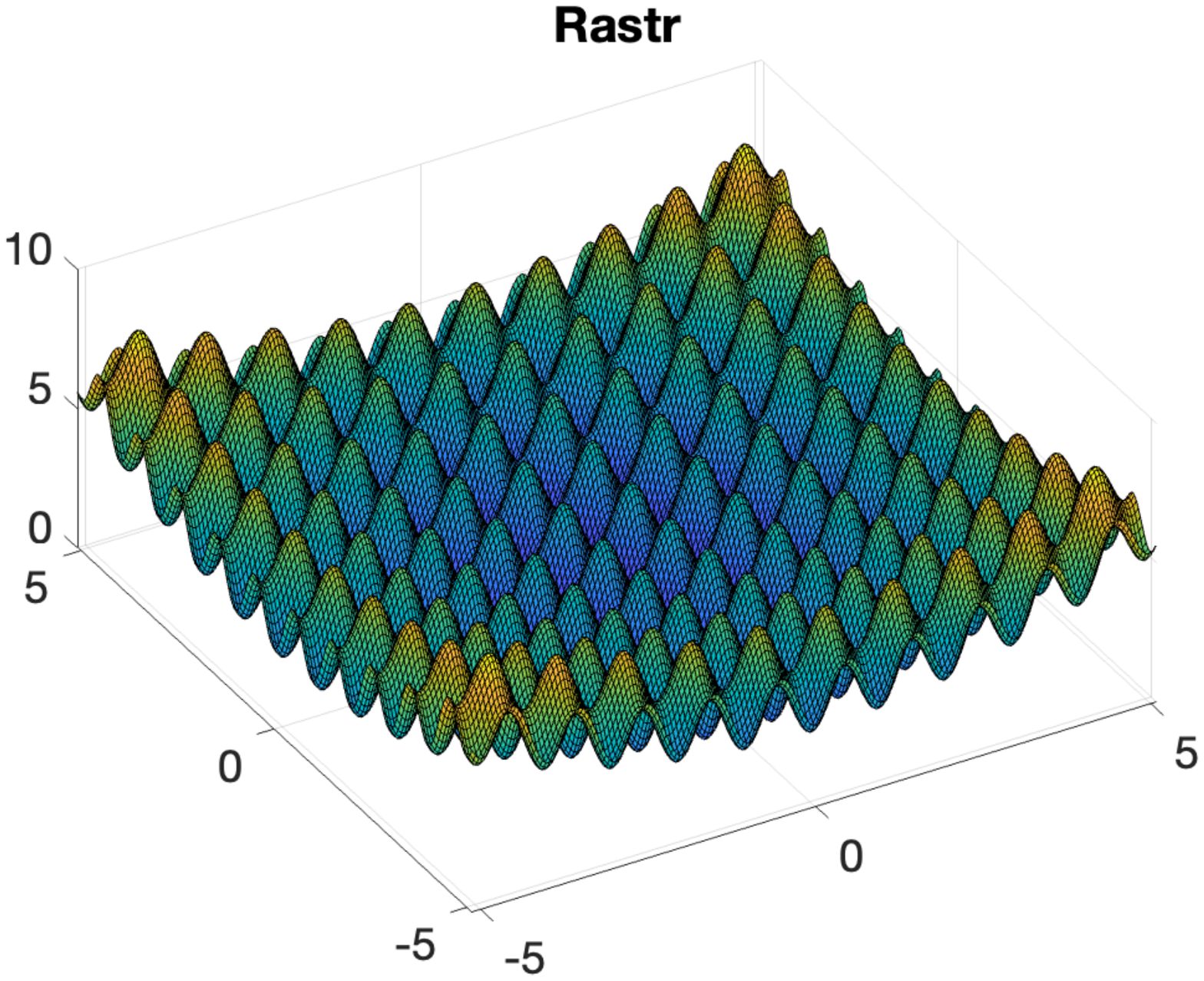}
    \caption{two-dimensional projections of ten-dimensional Schwefel (left) and Rastr (right)}
    \label{fig:test_func}
\end{figure}

In this section, we run our algorithms on the following test functions for prediction and Bayesian optimization:
\begin{align}
    &f_{\text{Schwefel}}(\BFx)=418.9829-\frac{1}{d}\sum_{d=1}^Dx_d\sin(\sqrt{\vert x_d\vert}),\quad \BFx\in(-500,500)^D,\label{eq:schwefel}\\
    &f_{\text{Rastr}}(\BFx)=10-\frac{1}{D}\sum_{d=1}^D\left(x_d^2-10\cos(2\pi x_d)\right) \BFx\in(-5.12,5.12)^D,\label{eq:rastr}
\end{align}
where \eqref{eq:schwefel} is the Schwefel function and \eqref{eq:rastr} is the Rastr function. Both of them are complex functions with many local minima and, hence, ideal test functions for salable data. Figure \ref{fig:test_func} shows the two-dimensional projections of our test functions. All the observations are corrupted by a standard normal noise, i.e., $\BFy=f(\BFx)+\varepsilon$, $\varepsilon\sim\mathcal{N}(0,1)$, for all $\BFx$.

We use Mat\'ern-$\frac{1}{2}$ kernel for the covariance of GP:
\[k(
\BFx,\BFx'
)=\sum_{d=1}^D \exp\left(-\theta\vert x_d-x_d'\vert\right).\]
 In this case, matrix $\BFPhi$ is a diagonal matrix (with band width $0$) and matrix $\BFA$ is a tridiagonal matrix (with band width $3$) We first run our algorithm for prediction of ten and twenty dimensional Schwefel and Rastr functions conditioned on data of size up to $30000$. We then run our algorithm on GP-UCB for searching the global minimizer of  five and ten dimensional Schwefel  and Rastr functions with sampling budget up to $3000$. 

  All the experiments are implemented in \textsc{Matlab} (version 2018a) on a laptop computer with macOS, 3.3 GHz Intel Core i5 CPU, and 8 GB of RAM (2133 Mhz).
The \textsc{Matlab} codes of the competing approaches are all publicly available.

\subsection{Prediction}
\begin{figure}
    \centering
    \includegraphics[width=.32\linewidth]{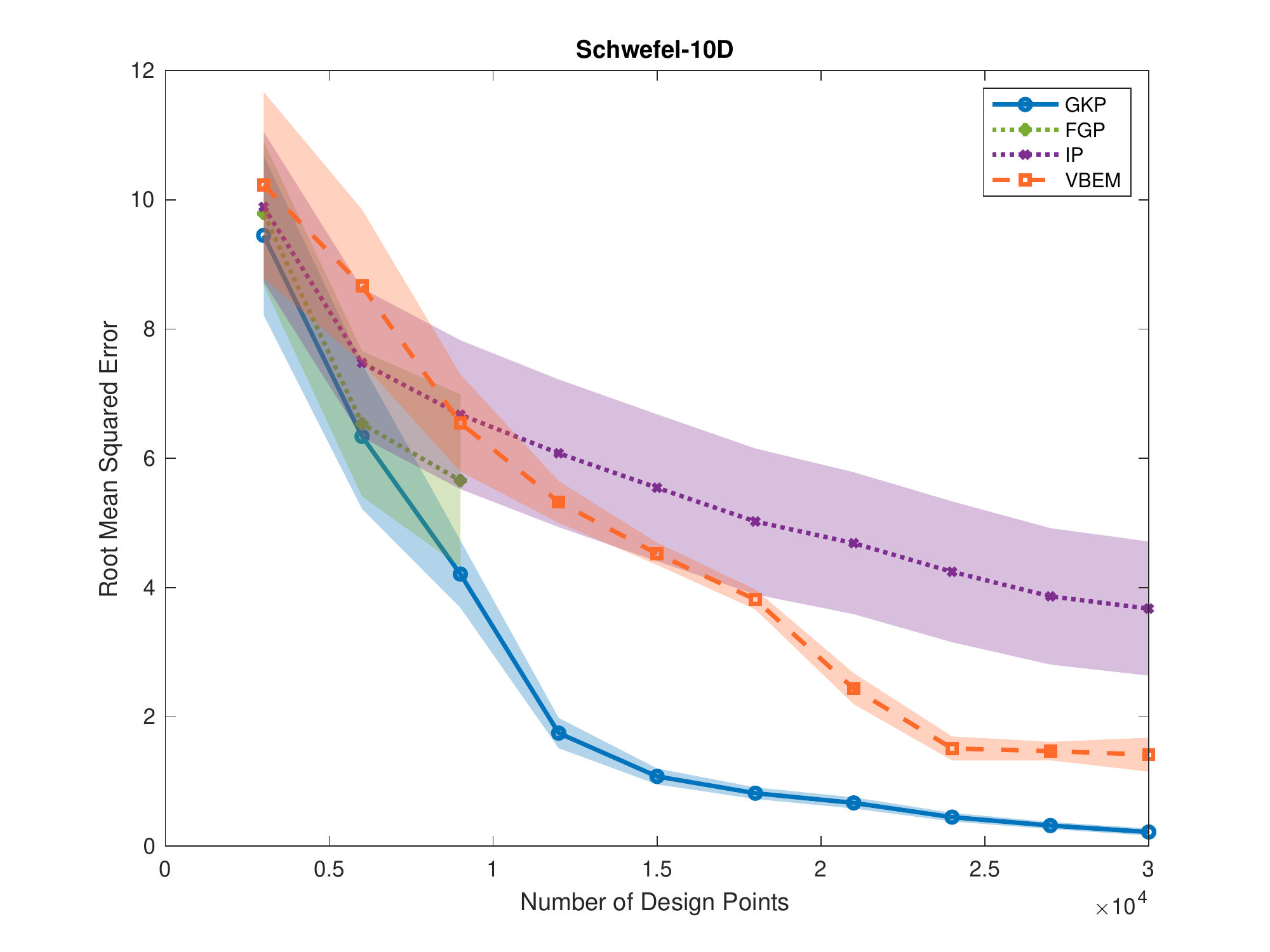}
    \includegraphics[width=.32\linewidth]{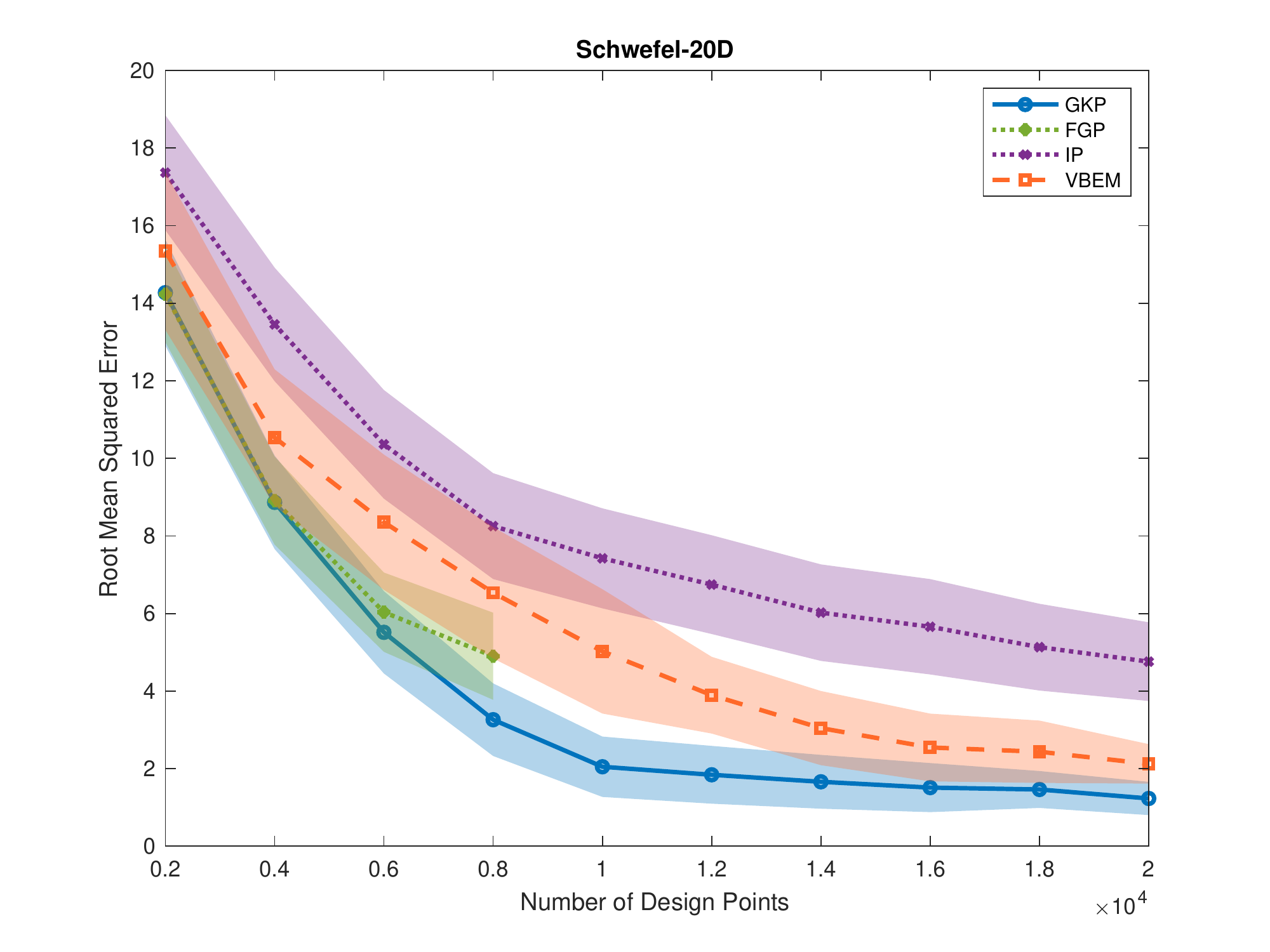}
    \includegraphics[width=.32\linewidth]{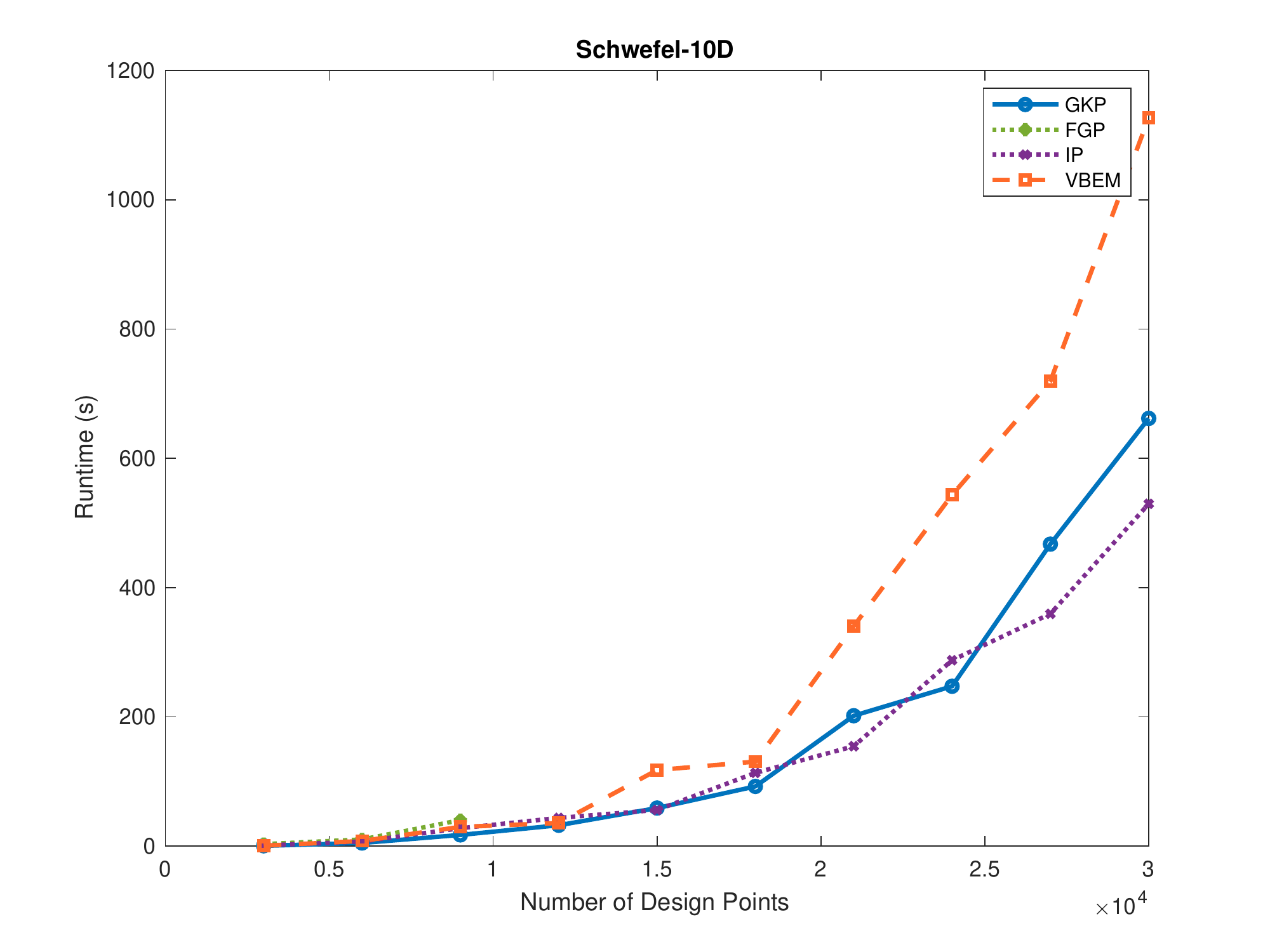}
    \includegraphics[width=.32\linewidth]{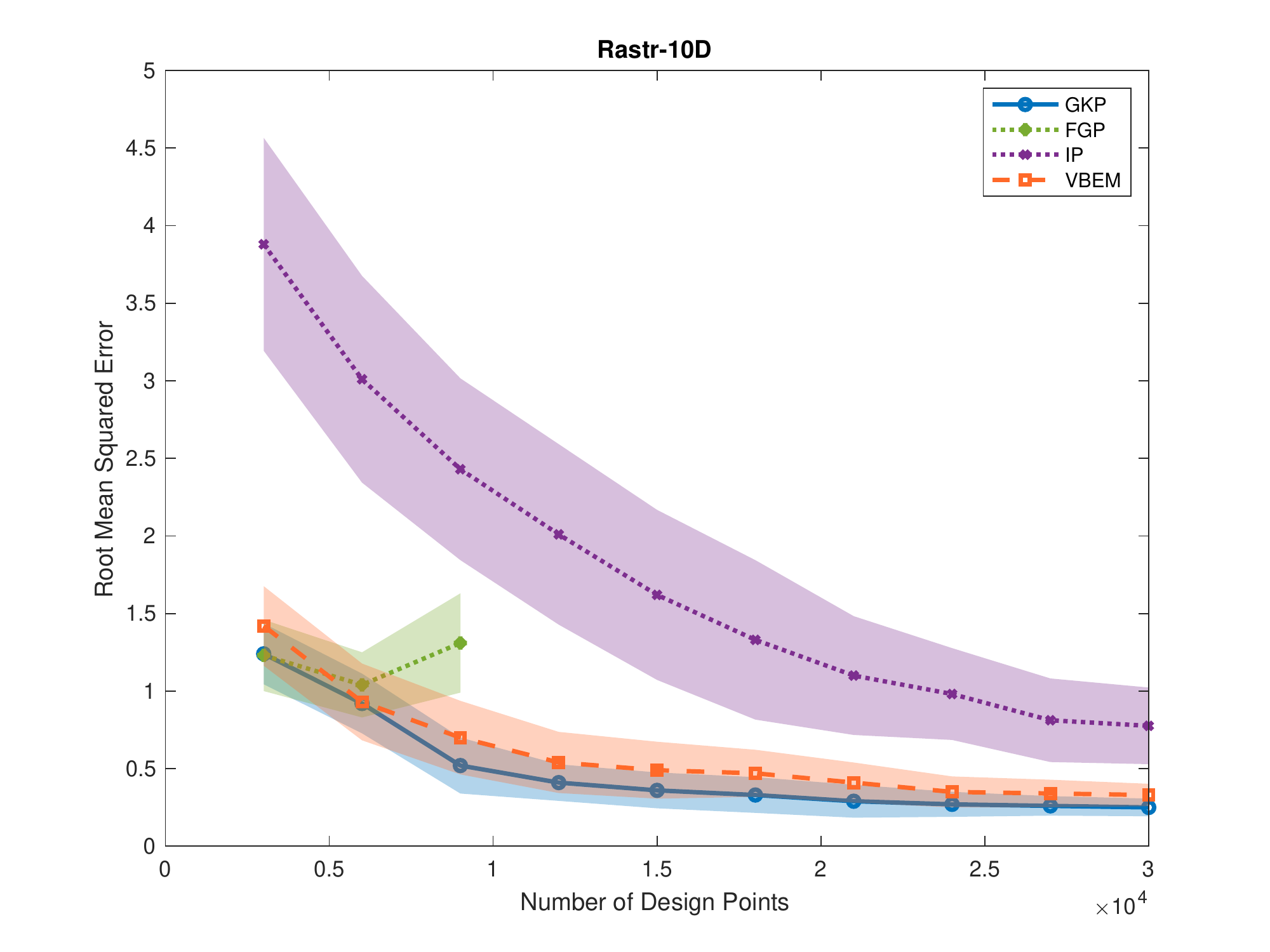}
    \includegraphics[width=.32\linewidth]{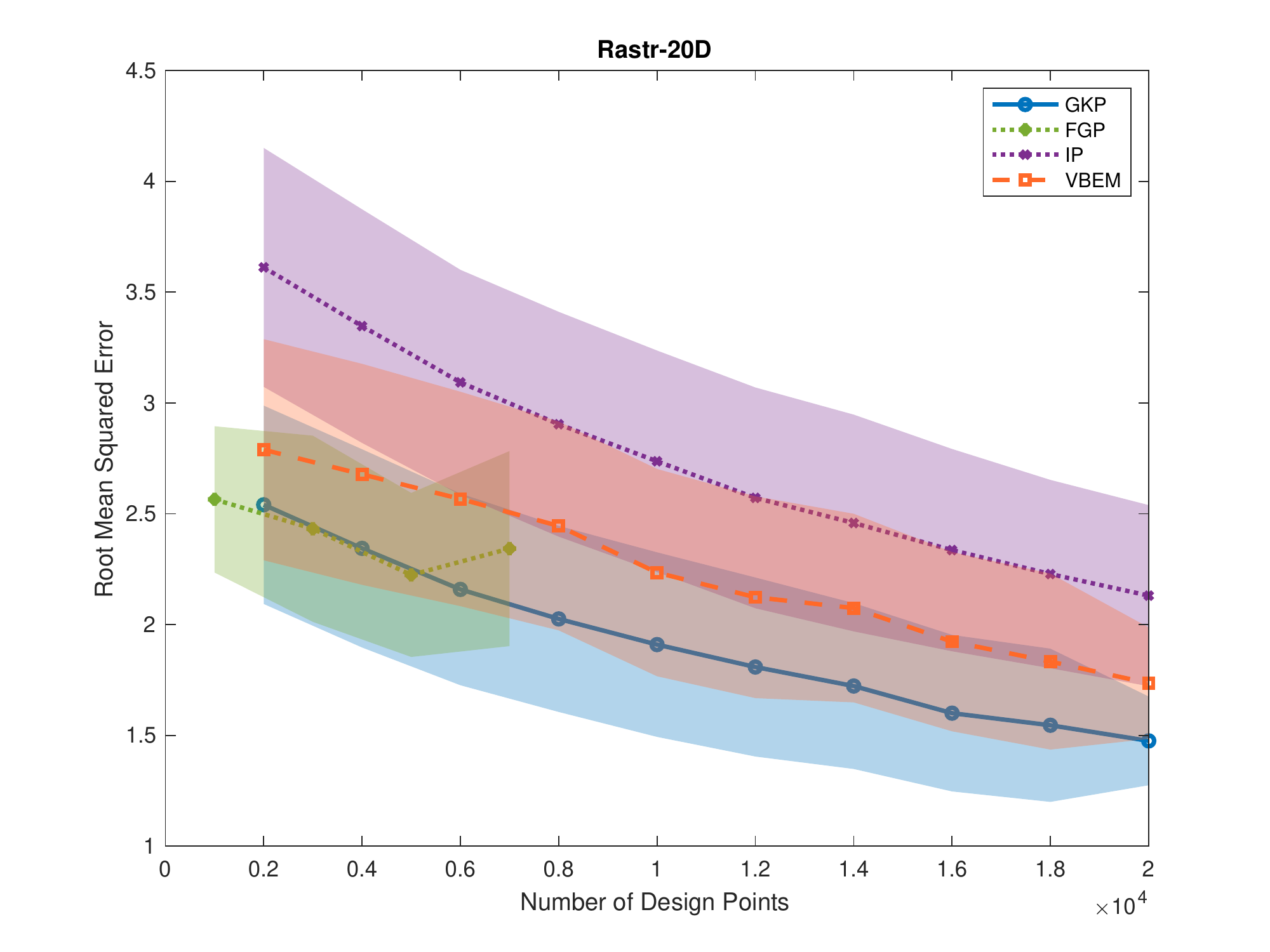}
    \includegraphics[width=.32\linewidth]{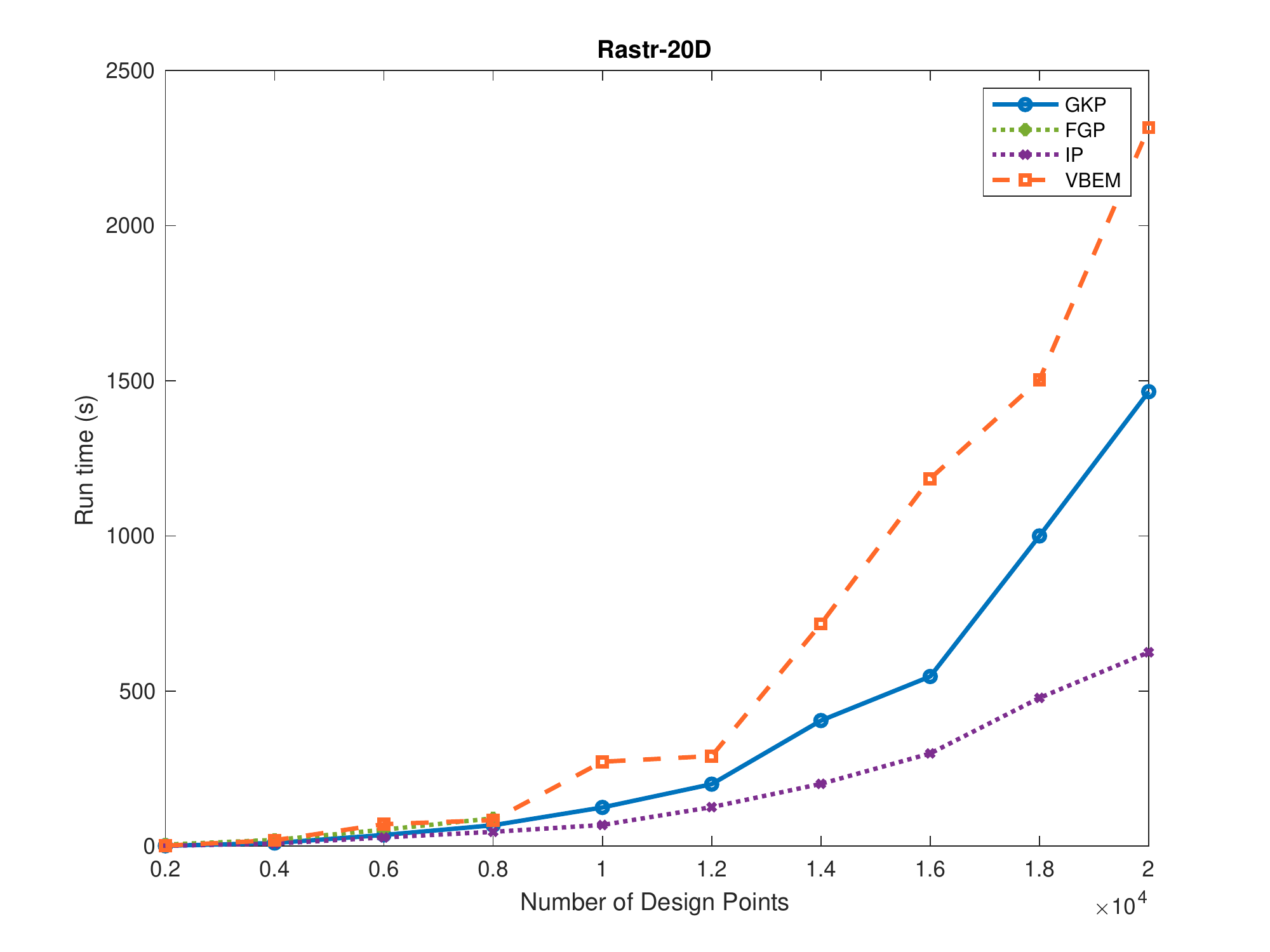}
    \caption{RMSE and computational time for test functions. The upper row corresponds to the Schwefel test function and the lower row corresponds to the Rastr test function. The three columns correspond to $d=10,20$ and computational time, respectively. The shaded areas areas represent standard deviation of the result}
    \label{fig:result}
\end{figure}

In this experiment, all the inputs $\BFx$ are uniformly generated from the domains of test function, i.e., $\BFx\sim\text{Unif}(-l,l)^D$ with $l=500,5.12$ and $D=10,20$. For dimension $D=10$, we examine our algorithm on all test functions with data size $n=3000,6000,\cdots,30000$.  For dimension $D=20$, we examine our algorithm on all test functions with data size $n=2000,4000,\cdots,20000.$ We use our algorithms to first compute the scale parameter that maximizes the log-likelihood: $\omega^*=\max_{\omega>0}l(\omega)$. We then use our method to compute the posterior mean $\hat{f}$  on 100 randomly selected test points $\{\BFx_i^*\}_{i=1}^{100}$ and compute the following Root Mean Squared Error (RMSE) to test the performance of our method:
\[\text{RMSE}=\sqrt{\frac{1}{100}\sum_{i=1}^{100}[\hat{f}(\BFx^*_i)-f(\BFx_i^*)]^2}\]
where $f$ is the true test function. We repeated the experiments 100 times and recorded the standard deviation of the RMSE to test the stability of each prediction model:
\[\text{STD}=\sqrt{\frac{1}{100}\sum_{i=1}^{100}(\text{RMSE}-\overline{\text{RMSE}})^2}\]
where $\overline{\text{RMSE}}$ is the averaged RMSE over the 100 macro repetition. Averaged computational time for computing the MLE and prediction is also recorded.
%we first run our algorithm to compute the log-likelihood values on a  discrete $\Theta$ set for the scale parameter $\theta$ and select $\theta^*=arg\max_{\theta\in\Theta}l(\theta)$ for the model. We then 
The following algorithms are used as benchmark:
\begin{enumerate}
    \item \textbf{Full GP} (\texttt{FGP}) \citep{RasmussenWilliams06}: naive implementation of GPs using \textit{GPML} tool box. 
    \item \textbf{Variational-Bayesian Expectation Maximization} (\texttt{VBEM}) \citep{gilboa2013scaling}: a variational inference approach for approximation the log-likelihood and posterior variance of additive GP;
    \item \textbf{Inducing Points} (\texttt{IP}): algorithm provided in the \textit{GPML} tool box. The number of inducing points $m$ is set as $m=\sqrt{n}$, which is the choice to achieve the optimal approximation power for Mat\'ern-$1/2$ correlation according to \cite{burt2019rates}.
\end{enumerate}
We call our method generalized Kernel packet (\texttt{GKP}). The experimental results are shown in Figure \ref{fig:result}.
First, our approach \texttt{GKP}, in general, outperforms the others significantly in prediction accuracy. 
Not only does 
\texttt{GKP} yield the lowest ${\text{RMSE}}$ in almost all the cases---regardless of the test function, dimensionality, or sample size---that are considered,
but the corresponding STDs of ${\text{RMSE}}$ also have the  smallest widths. 
Because the STDs are calculated via multiple macro-replications in each of which a random set of prediction points are sampled, 
this suggests that the predictions given by \texttt{GKP} are more stable than the competing approaches.

Second, the right column in Figure~\ref{fig:result}  
shows the average computational time of each approach in predicting the two test functions in different dimensions.
Compared with the two approaches that are not based on matrix approximations (i.e., \texttt{IP} and \texttt{VBEM}) the advantage of \texttt{GKP} is clear, especially when the data size is large.  However, compared with \texttt{IP}, the computational efficiency of \texttt{GKP} is lower but only by a small margin, especially when the sample size is large. 
This is not surprising because \texttt{IP} exploits low-rank approximations to accelerate matrix inversion.  
However, the acceleration in computation is achieved at the cost of prediction accuracy. 
Indeed, the $ {\text{RMSE}}$ associated with \texttt{IP} is markedly lower than that  associated with both \texttt{GKP} and \texttt{VBEM} in almost all cases. 
In a nutshell, \texttt{GKP} achieves a much higher prediction accuracy with a slightly lower computational efficiency than the two approximation approaches.

\subsection{Bayesion Optimization}
\begin{figure}
    \centering
    \includegraphics[width=.32\linewidth]{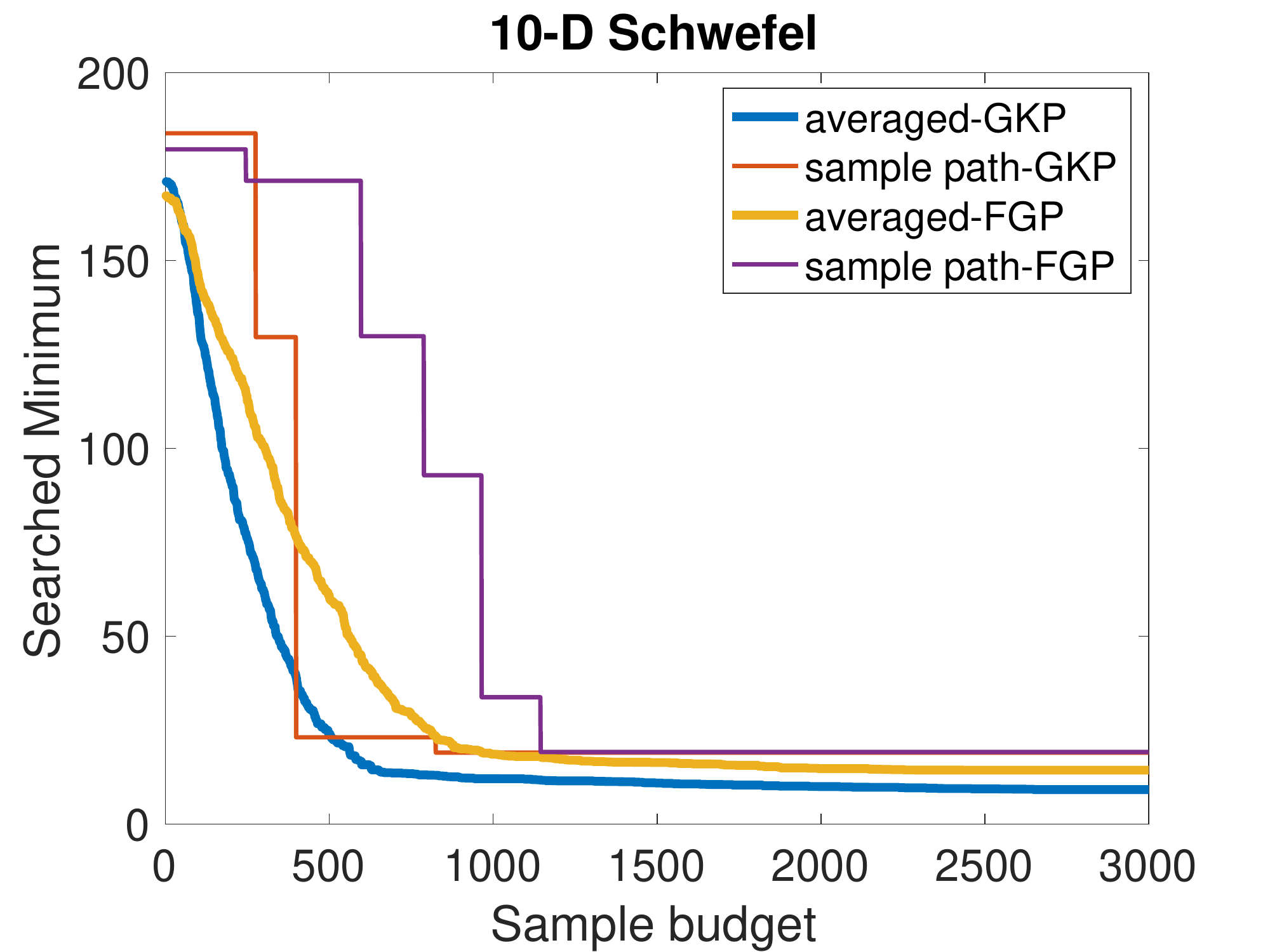}
    \includegraphics[width=.32\linewidth]{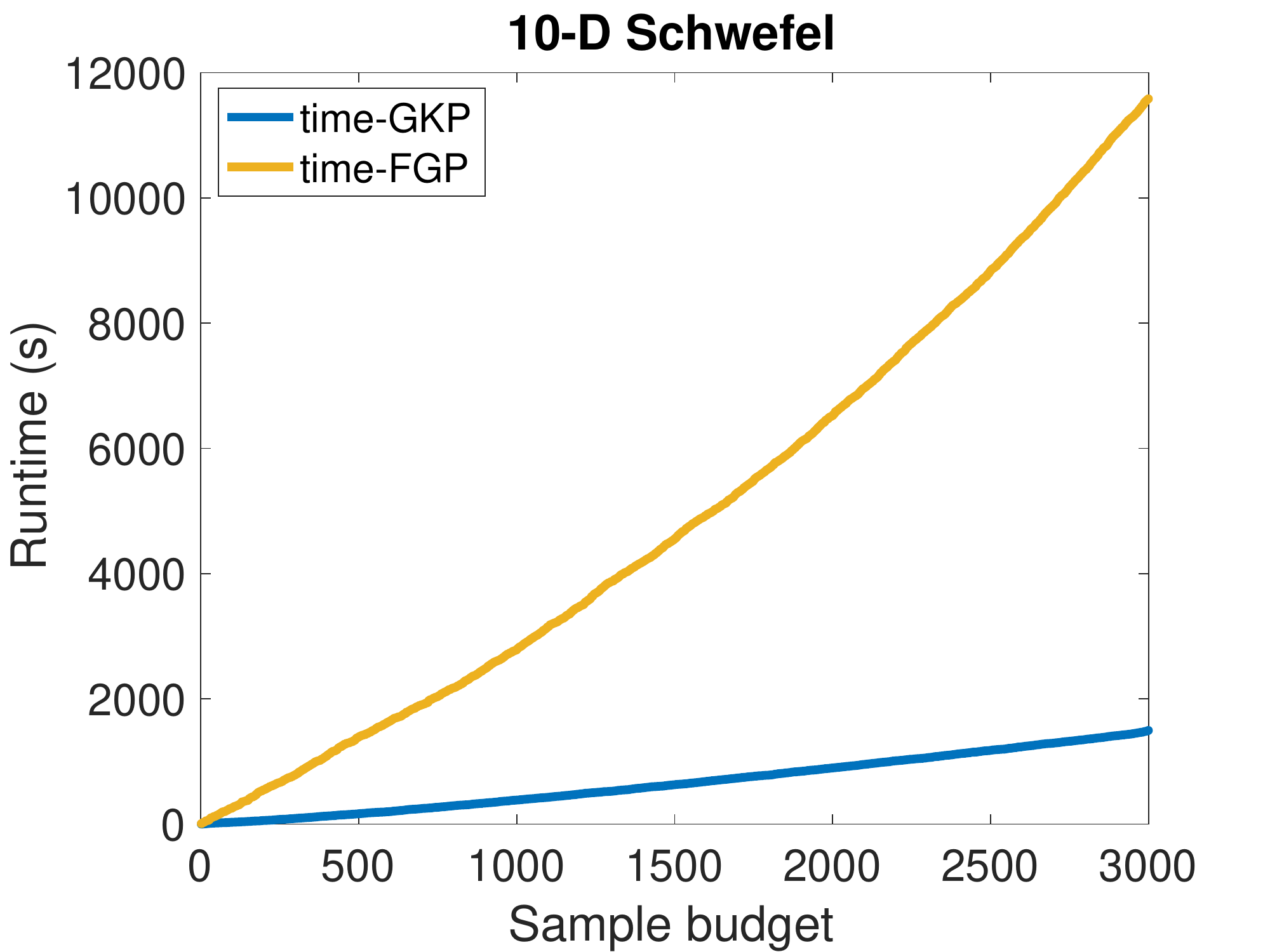}
    \includegraphics[width=.32\linewidth]{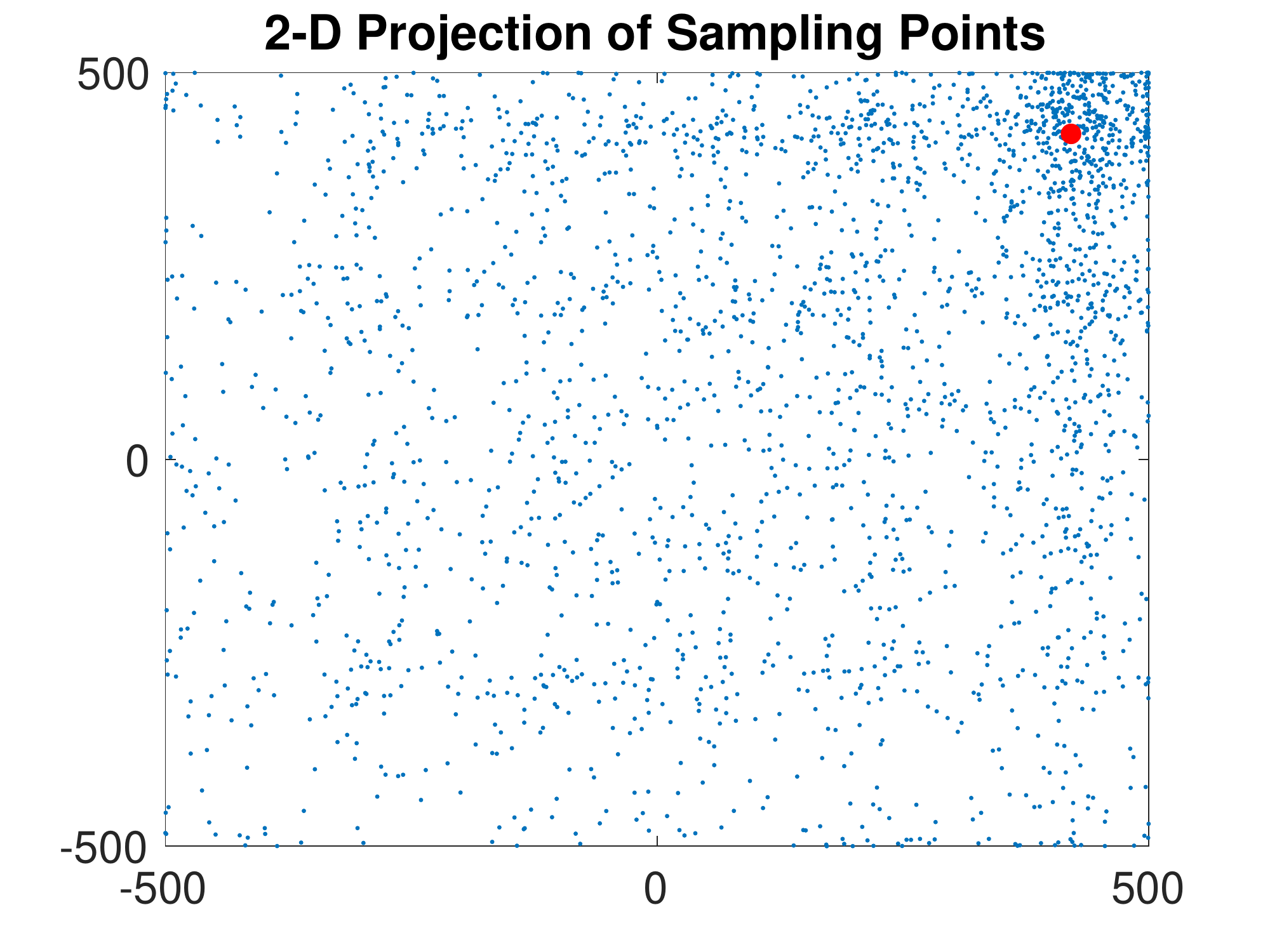}
    \includegraphics[width=.32\linewidth]{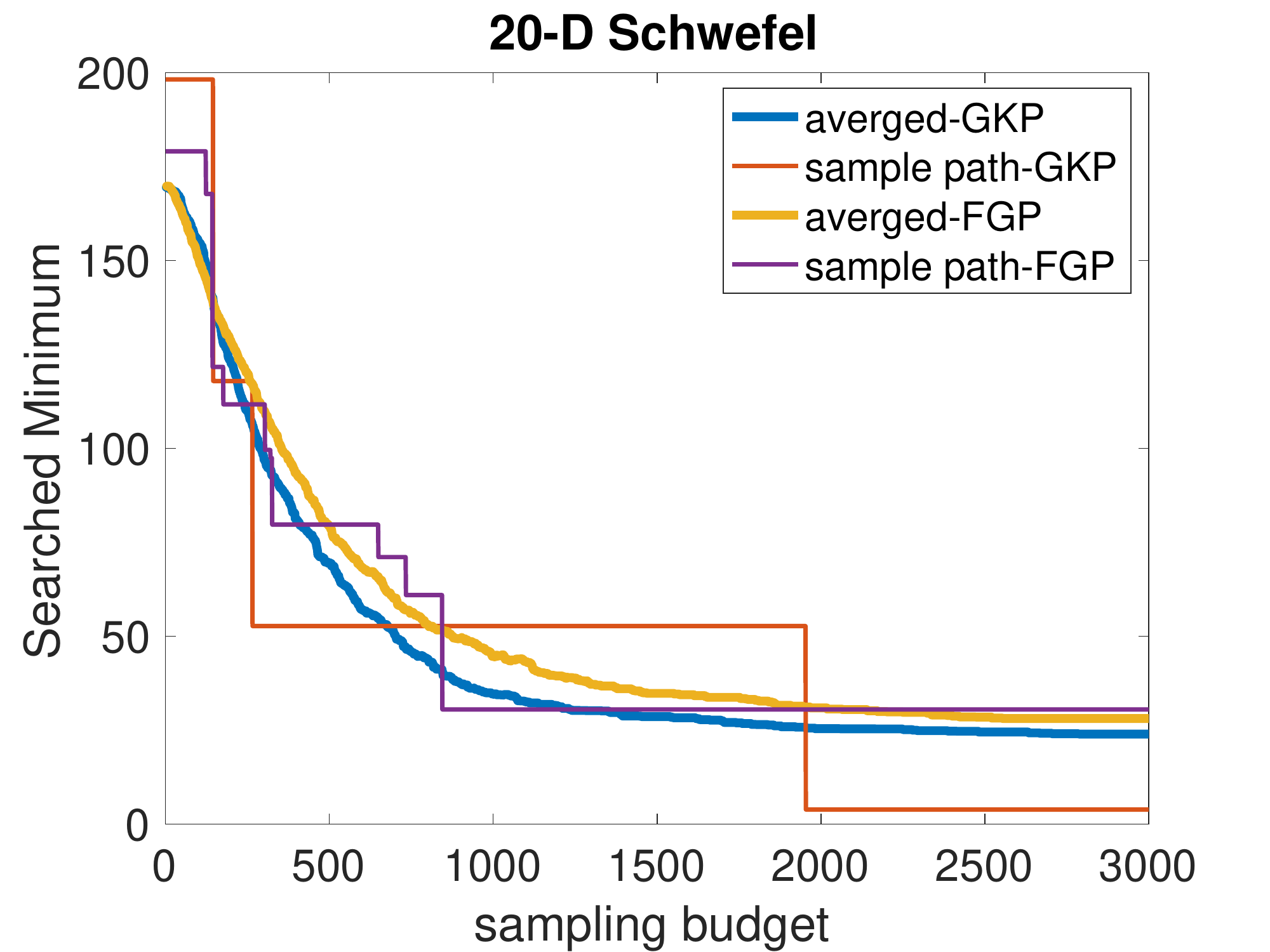}
    \includegraphics[width=.32\linewidth]{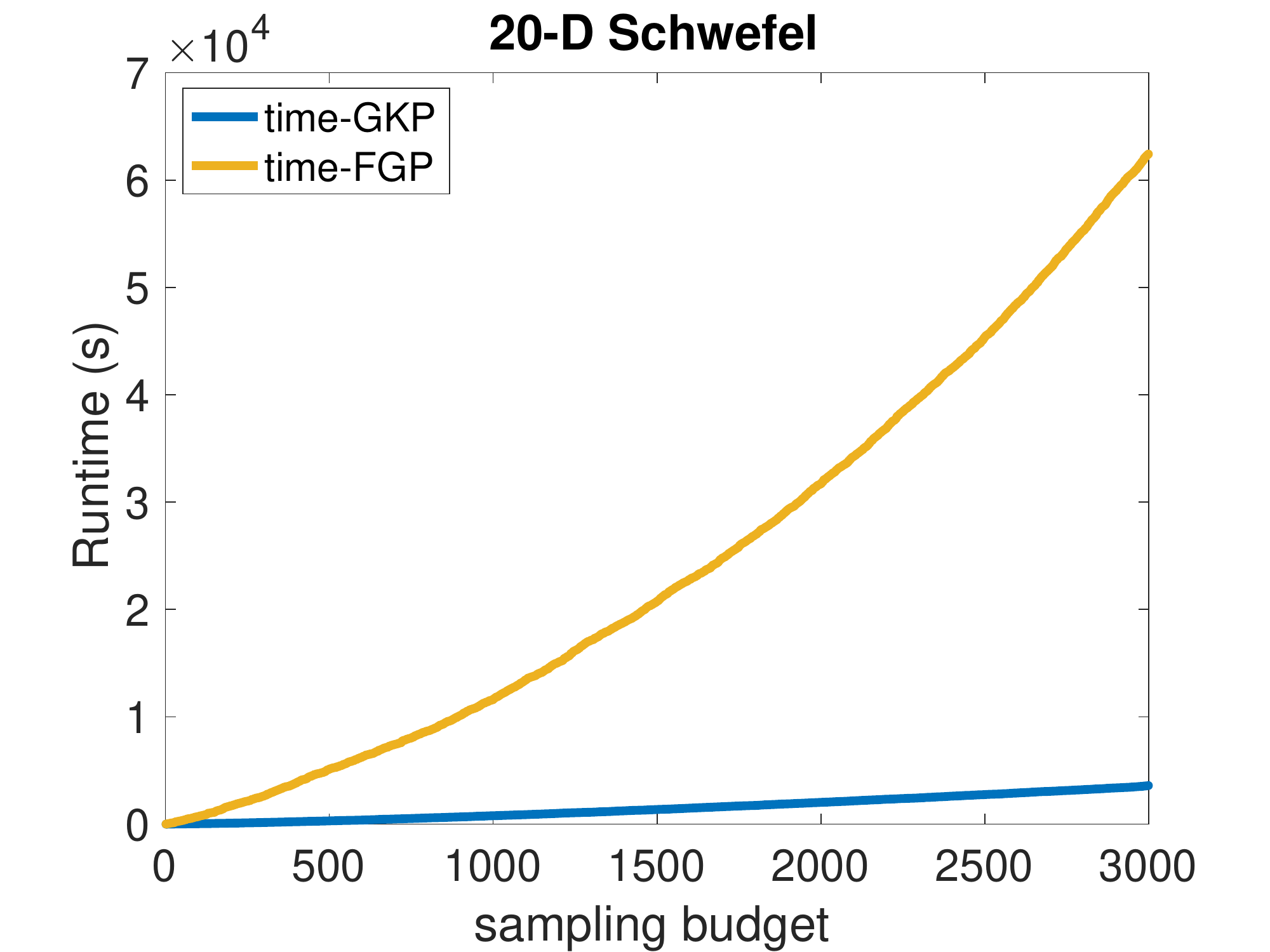}
    \includegraphics[width=.32\linewidth]{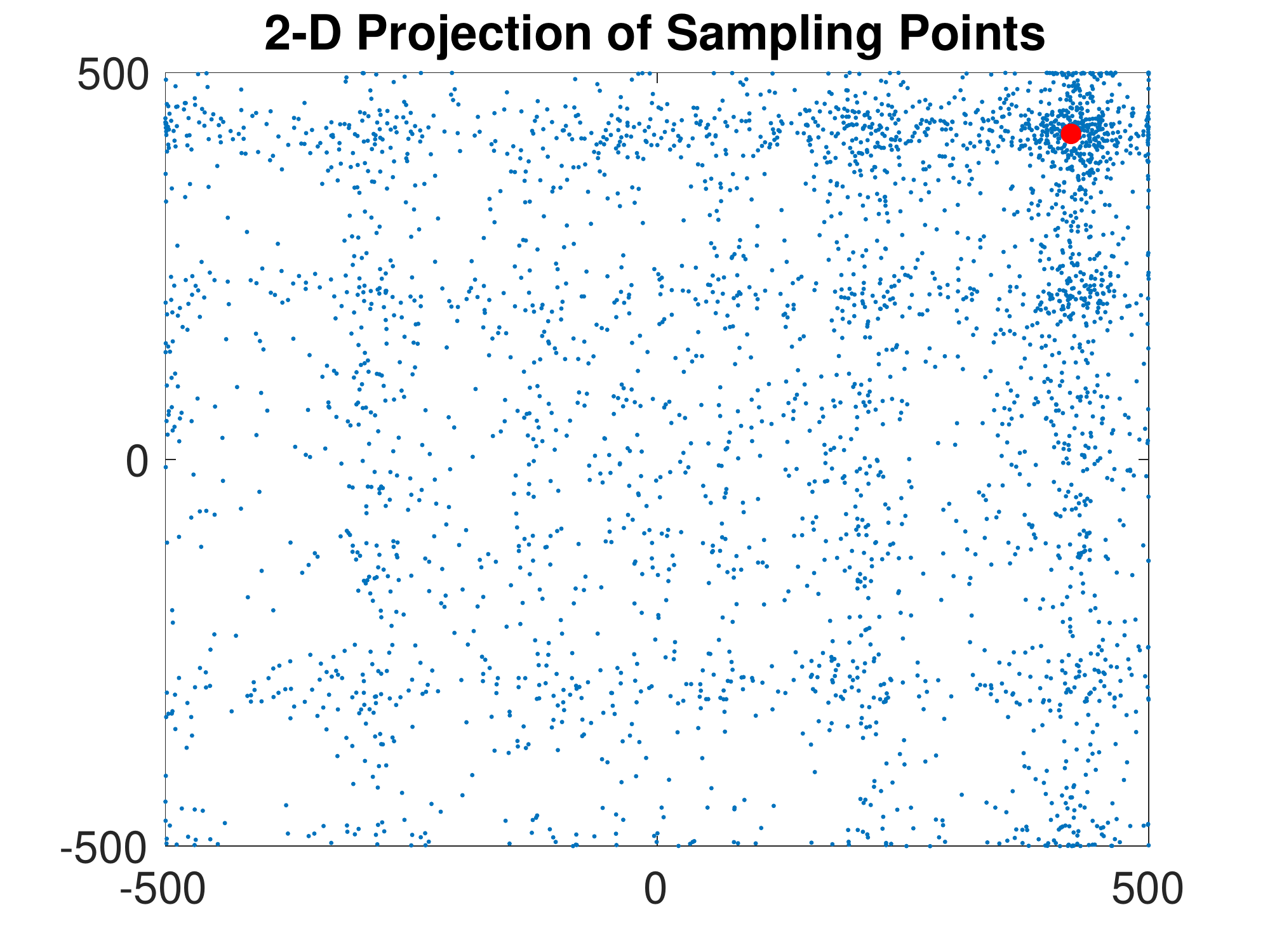}
    \caption{Searched Minimum, computational time, and sampling points. The upper row corresponds to the 10-dimensional Schwefel test function and the lower row corresponds to the 20-dimensional Schwefel test function. The left column is the minimum estimated by algorithms, the mideele is the computational times, the right column is the samples by \texttt{GKP}. }
    \label{fig:result_BO}
\end{figure}

In this experiment, we first randomly collect 100 sample points for the warm-up stage of Bayesian optimization algorithm. Then we use our algorithm \texttt{GKP} to compute the GP-UCB acquisition function for sequential samplings. At each sampling, we report the estimated minimizer of our algorithm. As a benchmark, we use full GP (FGP) to naively implement the GP-UCN algorithm.  We set the search spaces as $(-500,500)^D$ with $D=10$ or $20$. In each iteration, we first learn the hyperparameter of the additive GP conditioned on current data. We then use our algorithms to search the point that maximizes the acquisition function of GP-UCB. At the end of each iteration, we samply noisy value at a point that maximized current acquisition function. For $D=10$, we set the simulation budget to be $N=300, 6000, \cdots, 30000$. For $D=20$, we set the simulation budget to be $N=2000, 2000, \cdots, 20000$. Note that both functions are minimized at $(420.9687, 420.9687,\cdots,420.9687)$ and we have labeled the 2-D projection of the maximizer in the right column of \ref{fig:result_BO}.

The results of our Bayesian optimization experiments are presented in Figure \ref{fig:result_BO}. Our algorithm, \texttt{GKP}, is shown to have higher computational accuracy due to its efficiency and sparsity. Moreover, our algorithm requires much lower time complexity. The left column of Figure \ref{fig:result_BO} shows that \texttt{GKP} takes fewer iterations to estimate the minimizer, which, combined with the computational efficiency achieved by the sparsity of \texttt{GKP}, leads to its superior performance over \texttt{FGP}. As the data size increases during the sampling process, the advantage of \texttt{GKP} over \texttt{FGP} becomes even more evident. Finally, the right column of Figure \ref{fig:result_BO} shows that \texttt{GKP} spends a large portion of the sampling budget around the true  minimizer, which demonstrates the computational accuracy of our algorithm.

\section{Conclusion}

We present a novel approach to efficiently compute the posterior of an additive GP by decomposing it into a combination of one-dimensional GPs. Specifically, we leverage a recent development in sparse representation of one-dimensional Mat\'ern GPs to represent the posterior of an additive GP as a formulation by sparse matrices. Our approach allows us to compute the posterior in $\CalO(n)$ time and, given the posterior, to compute any Bayesian optimization acquisition function and its associated gradient in $\CalO(\log n)$ or even $\CalO(1)$ time. We evaluate the performance of our algorithms on complex test functions with scalable data and demonstrate their effectiveness.

The current study can be extended in several ways, the first being the inference of additive GPs with sparse additive terms. While we have assumed a full additive model in our paper, some additive models (e.g., \cite{raskutti2012minimax,cai2022stochastic}) assume that there are only a small number of unknown effective additive terms. Therefore, our study can be extended to efficiently compute the inference of these effective terms. Secondly, it is worth noting that many current deep learning models can be viewed as compositions of additive models. As such, our proposed algorithms could be applied to efficient inference of these models, including deep Gaussian processes \cite{damianou2013deep} and Bayesian neural networks \cite{mackay1995probable,neal2012bayesian,blundell2015weight}.  Finally, our current algorithms are only for additive GPs with Mat\'ern covariances but we believe that our algorithms can be generalized to other commonly used covariances, such as integrated Brownian motion \citep{SalemiStaumNelson19}, tensor Markov GPs \cite{ding2022sample}, and smoothing spline \citep{kimeldorf1970correspondence,kim2004smoothing}.

%%REFERENCES%%
%%%%%%%%%%%%%%%%%%%%%%%%%%%%%%%%%%%%%%%%%%%%%%%%%%%%%%%%%%%%%%%%%%%%%%%%%%%%%%%%%%%%%%%%%%%%%%%%%%%%%%%%%%%%%%%%%%%%%%%%%%%%%%%%%%%%
%% This template complies references using bibtex. You will need to use pomsref.bst file for biblography style.
%REFERENCES USING BIBTEX FILES
%%%%%%%%%%%%%%%%%%%%%%%%%%%%%%%%%%%%%%%%%%%%%%%%%%%%%%%%%%%%%%%%%%%%%%%%%%%%%%%%%%%%%%%%%%%%%%%%%%%%%%%%%%%%%%%%%%%%%%%%%%%%%%%%%%%%

\bibliographystyle{pomsref}

 \let\oldbibliography\thebibliography
 \renewcommand{\thebibliography}[1]{%
    \oldbibliography{#1}%
    \baselineskip14pt %Change this for line spacing within the same reference
    \setlength{\itemsep}{10pt}% %Change this for spacing between two referneces
 }
\bibliography{ref1}
%%%%%%%%%%%%%%%%%%%%%%%%%%%%%%%%%%%%%%%%%%%%%%%%%%%%%%%%%%%%%%%%%%%%%%%%%%%%%%%%%%%%%%%%%%%%%%%%%%%%%%%%%%%%%%%%%%%%%%%%%%%%%%%%%%%%

%Hayes, R. H., G. P. Pisano. 1996. Manufacturing strategy: At the intersection of two paradigm shifts. Production and Operations Management, 5 (1), 25-41.

%%%%%%%%%%%%%%%%%%%%%%%%%%%%%%%%%%%%%%%%%%%%%%%%%%%%%%%%%%%%%%%%%%%%%%%%%%%%%%%%%%%%%%%%%%%%%%%%%%%%%%%%%%%%%%%%%%%%%%%%%%%%%%%%%%%%
%% %If you don't use BiBTex, you can manually itemize references as shown in the referneces for the electronic comapanion. See below.
 %%%%%%%%%%%%%%%%%%%%%%%%%%%%%%%%%%%%%%%%%%%%%%%%%%%%%%%%%%%%%%%%%%%%%%%%%%%%%%%%%%%%%%%%%%%%%%%%%%%%%%%%%%%%%%%%%%%%%%%%%%%%%%%%%%%%

%% Here starts the e-companion (EC). Place your appendix content here.
%%%%%%%%%%%%%%%%%%%%%%%%%%%%%%%%%%%%%%%%%%%%%%%%%%%%%%%%%%
%\ECSwitch % Comment this line out if you do not need e-companion.
%%%%%%%%%%%%%%%%%%%%%%%%%%%%%%%%%%%%%%%%%%%%%%%%%%%%%%%%%%

%%% Main head for the e-companion

%If you don't use BiBTex, you can manually itemize references as shown below.

% Use this for adding Appendix
% Options are (1) APPENDIX (with or without general title) or
%             (2) APPENDICES (if it has more than one unrelated sections)
% Outcomment the appropriate case if necessary
%
\begin{APPENDIX}{}

\section{Proof of Theorem \ref{thm:posterior_block}}
\proof{Proof.}
The intuition of \eqref{eq:GP_posterior} relies on representing the conditional distribution $\pr(\CalG(\BFx^*)\vert\BFX,\BFY)$  in the following form:
\begin{align*}
    &\quad \pr\left(\CalG(\BFx^*)\vert\BFX,\BFY\right)\\
    &=\int\pr\left(\sum_{d=1}^D\CalG_d(x^*_D)\bigg\vert\CalG_1(\BFX_1),\cdots,\CalG_D(\BFX_D)\right)\nonumber\\
    &\quad\cdot\pr\left(\CalG_1(\BFX_1),\cdots,\CalG_D(\BFX_D)\bigg\vert\BFY\right)d\CalG_1(\BFX_1)\cdots d\CalG_D(\BFX_D)\\
    &\propto \int \pr\left(\sum_{d=1}^D\CalG_d(x^*_D)\bigg\vert\CalG_1(\BFX_1),\cdots,\CalG_D(\BFX_D)\right)\\
    &\quad\cdot\pr\left(\BFY \vert\CalG_1(\BFX_1),\cdots,\CalG_D(\BFX_D)\right)\pr\left(\CalG_1(\BFX_1),\cdots,\CalG_D(\BFX_D)\right) d\CalG_1(\BFX_1)\cdots d\CalG_D(\BFX_D)
\end{align*}
where the first equality is from the formula of  marginal distribution and the proportion relation is from Bayes rule. Remind that the $\CalG_d(\BFX_d)$ above should be treated as a vector indexed by $\BFX_d$ instead of values of $\CalG_d$ on $\BFX_d$. To directly prove the theorem using the above identity involves long calculations. Therefore, we use algebraic calculations to show that the posterior variance in \eqref{eq:posterior_block} is equivalent to the one in \eqref{eq:GP_posterior}. Proof for the posterior mean is similarly. 

Without loss of generality, assume $\sigma_y=1$, then
\begin{align}
    \quad &\boldsymbol{1}^T\boldsymbol{\gamma}_{\BFx^*}^T\BFK^{-1}[\BFK^{-1}+\BFS\BFS^T]^{-1}\BFK^{-1}\boldsymbol{\gamma}_{\BFx^*}\boldsymbol{1}\nonumber\\
    =&\boldsymbol{1}^T\boldsymbol{\gamma}_{\BFx^*}^T\BFK^{-1}\left(\BFK-\BFK\BFS[\rmI_n+\BFS^T\BFK\BFS]^{-1}\BFS^T\BFK\right)\BFK^{-1}\boldsymbol{\gamma}_{\BFx^*}\boldsymbol{1}\nonumber\\
    =& \boldsymbol{1}^T\boldsymbol{\gamma_{\BFx^*}}^T\BFK^{-1}\boldsymbol{\gamma_{\BFx^*}}\boldsymbol{1}-\boldsymbol{1}^T\boldsymbol{\gamma}_{\BFx^*}^T\BFS[\rmI_n+\BFS^T\BFK\BFS]^{-1}\BFS^T\boldsymbol{\gamma}_{\BFx^*}\boldsymbol{1}\label{eq:woodbury}
%=&\boldsymbol{1}^T\boldsymbol{\gamma_{\BFx^*}}^T\BFK^{-1}\boldsymbol{\gamma_{\BFx^*}}\boldsymbol{1}-\left[\sum_{d=1}^Dk_d(x^*_d,\BFX_d)\right]\left[\sum_{d=1}^Dk_d(\BFX_d,\BFX_d)+\rmI_n\right]^{-1}\left[\sum_{d=1}^Dk_d(\BFX_d,x^*_d)\right]\label{eq:woodbury}.
\end{align}
where
\begin{align*}
&[\rmI_n+\BFS^T\BFK\BFS]^{-1}=\left[\sum_{d=1}^Dk_d(\BFX_d,\BFX_d)+\rmI_n\right]^{-1}\\
&\BFS^T\boldsymbol{\gamma}_{\BFx^*}\boldsymbol{1}=\sum_{d=1}^Dk_d(\BFX_d,x^*_d).
\end{align*}
Substitute \eqref{eq:woodbury} into \eqref{eq:posterior_block}, it is straightforward to check that \eqref{eq:posterior_block} and \eqref{eq:GP_posterior} are equivalent.
\endproof

\section{Proof of Theorem \ref{thm:add_likelihood}}
\proof{Proof.}
    For \eqref{eq:add_likelihood}, we can see in \eqref{eq:loglik} that $l$ consists of the quadratic term and the log determinant term. For the quadratic term, we first use the following identity:
    \begin{align}
        &\quad\frac{\BFY}{D}^T\BFS^T\BFK_{\BFthete}^{-1}\left[\BFK_{\BFthete}^{-1}+\sigma_y^{-2}\BFS\BFS^T\right]^{-1}\BFK_{\BFthete}^{-1}\BFS\frac{\BFY}{D}\nonumber\\
        &= \frac{\BFY}{D}^T\BFS^T\BFK_{\BFthete}^{-1}\left(\BFK_{\BFthete}-\BFK_{\BFthete}\BFS\left[\BFS^T\BFK_{\BFthete}\BFS+\sigma_y^2\rmI_n\right]^{-1}\BFS^T\BFK_{\BFthete}\right)\BFK_{\BFthete}^{-1}\BFS\frac{\BFY}{D}\nonumber\\
        &= \frac{\BFY}{D}^T\BFS^T\BFK_{\BFthete}^{-1}\BFS\frac{\BFY}{D}-\frac{\BFY}{D}^T\BFS^T\BFS[\BFS^T\BFK_{\BFthete}\BFS+\sigma^2\rmI_n]^{-1}\BFS^T\BFS\frac{\BFY}{D} \label{eq:quadratic_idnetity}
    \end{align}
    where the second line is from Woodbury matrix identity. Notice that 
    \begin{equation}
        \frac{1}{D}\BFS^T\BFS\BFY=\frac{1}{D}\sum_{d=1}^D\BFY=\BFY\label{eq:loglik_Ssum}
    \end{equation}
    So the quadratic term $\BFY^T[k(\BFX,\BFX\vert\BFthete))+\sigma^2\rmI_n]^{-1}\BFY$ can be written as:
    \begin{align*}
        &\quad\BFY^T[k(\BFX,\BFX\vert\BFthete)+\sigma^2\rmI_n]^{-1}\BFY\nonumber\\&=\frac{\BFY}{D}^T\BFS^T\BFS[\BFS^T\BFK_{\BFthete}\BFS+\sigma^2\rmI_n]^{-1}\BFS^T\BFS\frac{\BFY}{D}\\&=\frac{\BFY}{D}^T\BFS^T\BFK_{\BFthete}^{-1}\BFS\frac{\BFY}{D}-\frac{\BFY}{D}^T\BFS^T\BFK_{\BFthete}^{-1}\left[\BFK_{\BFthete}^{-1}+\sigma_y^{-2}\BFS\BFS^T\right]^{-1}\BFK_{\BFthete}^{-1}\BFS\frac{\BFY}{D}
    \end{align*}
    where the second line is from \eqref{eq:loglik_Ssum} and the third line is from \eqref{eq:quadratic_idnetity}. The above equation gives the quadratic term in \eqref{eq:add_likelihood}.

    For the log determinant term, we can use the matrix determinant lemma:
    \begin{equation}
\vert\BFS^T\BFK_{\BFthete}\BFS+\sigma_y^2\rmI_n\vert=\vert\BFK_{\BFthete}^{-1}+\sigma_y^{-2}\BFS\BFS^T\vert\cdot\vert\BFK_{\BFthete}\vert\cdot\vert\sigma_y^2\rmI_n\vert\label{eq:mat_det_lem}
    \end{equation}
    which gives exactly the log determinant terms of \eqref{eq:add_likelihood}.

    For \eqref{eq:add_likelihood_gradient}, we can use matrix derivative rules and apply Woodbury matrix identity to directly get the result.
\endproof

\section{Generalized Kernel Packet}
The derivative of a Mat\'ern kernel with half-integer smoothness parameter $\nu$ is of the following form:
\begin{equation}\label{eq:matern_explicit}
    k(x,x'\vert\omega)=\sigma \exp({-{\omega}{\vert x-x'\vert}})\frac{q!}{2q!}\left(\sum_{l=0}^{q}\frac{(q+l)!}{l!(q-l)!}({2\omega}{\vert x-x'\vert})^{q-l}\right)
\end{equation}
where $q=\nu-\frac{1}{2}$. Without loss of generality, we let $\sigma=1$. We first present the following theorem as a generalized version of 1 in Theorem \ref{thm:KP}.

\begin{theorem}[Central ]\label{thm:generalized_KP}
   Let $k(\cdot,\cdot|\omega)$ be a Mat\'ern-$\nu$ kernel with half-integer $\nu$ and $q=\nu-1/2$. For any $p=2\nu+4$ points sorted in increasing order $\{x_i\}_{i=1}^p$,   let $(b_1,\cdots,b_p)$ be the solution of the following system of equations:
    \begin{equation}\label{eq:generalized_KP}
        \sum_{i=1}^{p} b_i x_i^l \exp(\delta \omega x_i)=0,
    \end{equation}
with $l=0,\ldots,q+1$, and $\delta=\pm 1$. Then function:
$\psi_{(x_1,\cdots,x_p)}=\sum_{i=1}^pb_i\partial_\omega k(\cdot,x_i|\omega)$
is non-zero only on interval $(x_1,x_p)$.
\end{theorem}
\proof{Proof.}    
For any $x<x_1<x_2<\cdots,x_p$, we have
\begin{align}
    \partial_\omega  k(x,x_i\vert\omega)&=\partial_\omega \left(\exp\left({{\omega} (x-x_i)}\right)\frac{q!}{2q!}\left(\sum_{l=0}^{q}\frac{(q+l)!}{l!(q-l)!}({2\omega}(x_i- x))^{q-l}\right)\right)\nonumber\\
    &=(x-x_i)\exp({{\omega} (x-x_i)})\frac{q!}{2q!}\left(\sum_{l=0}^{q}\frac{(q+l)!}{l!(q-l)!}({2\omega}(x_i- x))^{q-l}\right)\nonumber\\
    &\quad +\exp({{\omega} (x-x_i)})\frac{q!}{2q!}\left(\sum_{l=0}^{q-1}\frac{(q+l)!}{l!(q-l-1)!}(2x_i- 2x)^{q-l}\omega^{q-l-1}\right)\nonumber\\
    &=-\exp(\omega(x-x_i))\frac{q!}{2q!}\sum_{s=2}^{q}\frac{(2q-s)!}{(q-s+1)!(s-2)!}(2\omega)^{s-1}(x_i-x)^{s}\nonumber\\
    &\quad -\exp(\omega(x-x_i))\frac{q!2^q\omega^q}{2q!}(x_i-x)^{q+1}\label{eq:sum_k_derivative}
\end{align}
where the first summation in \eqref{eq:sum_k_derivative} equals $0$ if $q\leq 1$. We can only consider the case $q\geq2$. For $q\leq 1$, analysis i similar. When $q\geq2$, \eqref{eq:sum_k_derivative} can be unified as
\begin{align}
     \partial_\omega  k(x,x_i\vert\omega)&= -\exp(\omega(x-x_i))\frac{q!}{2q!}\sum_{s=2}^{q+1}\frac{(2q-s)!}{(q-s+1)!(s-2)!}(2\omega)^{s-1}(x_i-x)^{s}\nonumber\\
    &= -\exp(\omega(x-x_i))\frac{q!}{2q!}\sum_{s=2}^{q+1}\sum_{l=0}^s\frac{(2q-s)!}{(q-s+1)!(s-2)!}(2\omega)^{s-1}\frac{s!}{l!(s-l)!}x_i^l(-x)^{s-l}\nonumber\\
    &= -\exp(\omega(x-x_i))\frac{q!}{2q!}\sum_{t=0}^{q+1}x_i^l \sum_{s=\max\{l,2\}}^{q+1}\frac{(2q-s)!}{(q-s+1)!(s-2)!}(2\omega)^{s-1}\frac{s!}{l!(s-l)!} (-x)^{s-l}\nonumber\\
    &\coloneqq\sum_{l=0}^{q+1}x_i^l\exp(\omega(x-x_i))C(x,q,l)\label{eq:generalized_KP_2}
\end{align}
where $C(x,q,t)$ in \eqref{eq:generalized_KP_2} is independent of $x_i$. Then for for any $\{b_i\}_{i=1}^p$ satisfying \eqref{eq:generalized_KP}, we have
\begin{align}
    \psi_{(x_1,\cdots,x_p)}(x)&=\sum_{i=1}^pb_i \partial_\omega  k(x,x_i\vert\omega)\nonumber\\
    &=\sum_{i=1}^pb_i\sum_{l=0}^{q+1}x_i^l\exp(\omega(x-x_i))C(x,q,l)\nonumber\\
    &=\sum_{l=0}^{q+1}e^{\omega x}C(x,q,l)\sum_{i=1}^pb_ix_i^le^{-\omega x_i}=0\label{eq:generalized_KP_3}
\end{align}
where the second line is from \eqref{eq:generalized_KP_2} and the third line is from condition \eqref{eq:generalized_KP}.

Similarly, for any $x>x_p>\cdots>x_1$, we have the identity
\begin{align}
    \partial_\omega  k(x,x_i\vert\omega)&=\partial_\omega \left(\exp\left({{\omega} (x_i-x)}\right)\frac{q!}{2q!}\left(\sum_{l=0}^{q}\frac{(q+l)!}{l!(q-l)!}({2\omega}(x- x_i))^{q-l}\right)\right)\nonumber\\
    &=\sum_{l=0}^{q+1}x_i^l\exp(\omega(x_i-x))C'(x,q,l)
\end{align}
and the same calculation as \eqref{eq:generalized_KP_3} shows
\begin{align}
    \psi_{(x_1,\cdots,x_p)}(x)
    &=\sum_{l=0}^{q+1}e^{-\omega x}C'(x,q,l)\sum_{i=1}^pb_ix_i^le^{\omega x_i}=0.\label{eq:generalized_KP_4}
\end{align}
Putting \eqref{eq:generalized_KP_3} and \eqref{eq:generalized_KP_4} together, we can have the final result.
\endproof

The following Theorem is the generalization of 2 in Theorem \ref{thm:KP}.
\begin{theorem}[One-sided]
     Let $k(\cdot,\cdot|\omega)$ be a Mat\'ern-$\nu$ kernel with half-integer $\nu$ and $q=\nu-1/2$. For any $p$ points sorted in increasing order $\{x_i\}_{i=1}^p$ with $\nu+\frac{5}{2}\leq p<2\nu+4$, let $(b_1,\cdots,b_p)$ be the solution of the following system of equations
    \begin{equation}\label{eq:generalized_KP_one_side}
    \sum_{i=1}^{p} b_ix_i^{l} \exp\{h\omega x_i\}=0,\quad \sum_{i=1}^{p} b_i x_i^{r} \exp\{-h\omega x_i\}=0,
\end{equation}
where $l=0,\ldots,q+1$, and the second term comprises auxiliary equations with $r=0,\ldots,p-\nu-7/2$ (if $p-\nu-7/2<0$, skip the right side of \eqref{eq:generalized_KP_one_side}). If $h=1$, then function:
$\psi_{(x_1,\cdots,x_p)}=\sum_{i=1}^pb_i\partial_\omega k(\cdot,x_i|\omega)$
is non-zero only on interval $(-\infty,x_p)$; If $h=-1$, then function:
$\psi_{(x_1,\cdots,x_p)}=\sum_{i=1}^pb_i\partial_\omega k(\cdot,x_i|\omega)$
is non-zero only on interval $(x_1,\infty)$.\label{thm:generalized_KP_one_sided}
\end{theorem}
\proof{Proof.}
    We can use reasoning similar to the proof for Theorem \ref{thm:generalized_KP}. For any $x<x_1<\cdots<x_p$, according to \eqref{eq:generalized_KP_2}, 
    \[\partial_\omega k(x,x_i|\omega)=\sum_{l=0}^{q+1}x_i^l\exp(\omega(x-x_i))C(x,q,l).\]
    For any $\{b_i\}_{i=1}^p$ satisfying \eqref{eq:generalized_KP_one_side} with $h=-1$, we immediately have
    \[\psi_{(x_1,\cdots,x_p)}(x)=\sum_{i=1}^pb_i\partial_\omega k(x,x_i|\omega)=\sum_{l=0}^{q+1}e^{\omega x}C(x,q,l)\sum_{i=1}^pb_ix_i^le^{-\omega x_i}=0.\]
    For any $x>x_p>\cdots>x_1$, we only need to switch the sign of $x$ and $x_i$ to get the final result.
\endproof

Theorem \ref{thm:generalized_KP} and \ref{thm:generalized_KP_one_sided} are exactly the same as Theorem \ref{thm:KP} except that the coefficients $\{b_i\}_{i=1}^p$ for $\partial_\omega k(\cdot,\cdot\vert\omega) $ smoothness parameter $\nu$ are the coefficients $\{a_i\}_{i=1}^p$ for Mat\'ern-$\nu+1$ kernel packet with the same scale hyperparameter $\omega$. Using  Theorem \ref{thm:generalized_KP}, we can prove Theorem \ref{thm:banded_factor_gradient}.
\subsection{Proof of Theorem \ref{thm:banded_factor_gradient}}
\proof{Proof.}
    The fact that $\BFB$ is a $\nu+\frac{3}{2}$-banded matrix is a direct result of \cite[sec 3.1]{chen2022kernel} by treating $\BFB$ as the coefficient matrix $\BFA$ for Mat\'ern-$(\nu+1)$ KP. According to Algorithm \ref{alg:banded_factorization_gradient}, the $i$-th row of $\BFPsi$ for $1\leq i \leq \nu+\frac{3}{2}$ is
    \[[\BFPsi]_{i,j}=\sum_{s=1}^{i+\nu+\frac{3}{2}}\BFB_{i,s}\partial_\omega k(x_s,x_j\vert\omega)\]
    According to Theorem \ref{thm:generalized_KP_one_sided}, $[\BFPsi]_{i,j}=0$ for any $j\geq i+\nu+\frac{3}{2}$ because $[\BFPsi]_{i,j}$ is the value of a left-sided generalized KP associated to sorted points $\{x_j\}_{j=1}^{i+\nu+\frac{3}{2}}$.
    
    For $\nu+\frac{3}{2}<i<n-\nu-\frac{1}{2}$, the $i$-th row of $\BFPsi$ is 
    \[[\BFPsi]_{i,j}=\sum_{s=i-\nu-\frac{3}{2}}^{i+\nu+\frac{3}{2}}\BFB_{i,s}\partial_\omega k(x_s,x_j\vert\omega).\]
    According to Theorem \ref{thm:generalized_KP}, $[\BFPsi]_{i,j}=0$ for any $|j-i|\geq \nu+\frac{3}{2}$ because $[\BFPsi]_{i,j}$ is the value of a central generalized KP associated to sorted points $\{x_j\}_{i-\nu-\frac{3}{2}}^{i+\nu+\frac{3}{2}}$.

    For $i\geq n-\nu-\frac{1}{2}$, the $i$-th row of $\BFPsi$ is 
    \[[\BFPsi]_{i,j}=\sum_{s=i-\nu-\frac{3}{2}}^{n}\BFB_{i,s}\partial_\omega k(x_s,x_j\vert\omega).\]
    According to Theorem \ref{thm:generalized_KP_one_sided}, $[\BFPsi]_{i,j}=0$ for any $j\leq  i-\nu-\frac{3}{2}$ because $[\BFPsi]_{i,j}$ is the value of a right-sided generalized KP associated to sorted points $\{x_j\}_{j=i-\nu-\frac{3}{2}}^{n}$.

    To summarise, $[\BFPsi]_{i,j}=0$ for any $\vert j-i\vert>\nu+\frac{1}{2}$ so it is a $(\nu+\frac{1}{2})$-banded matrix.

    To prove the invertibility of $\BFB$, we can use \cite[Theorem 13 ]{chen2022kernel}, which states that the Gram matrix $\BFB \BFK$ is of full rank where $\BFK$ is the covariance matrix induced by Matern-$(\nu+1)$ kernel and any non-overlapped sorted points $\{x_i\}_{i=1}^n$. Therefore, $\BFB$ must bt invertible.
\endproof

\end{APPENDIX}
%
%   or
%

%%%%%%%%%%%%%%%%%
\end{document}